\documentclass[10pt,twocolumn,letterpaper]{article}

\usepackage[pagenumbers]{cvpr}      
\makeatletter
\@namedef{ver@everyshi.sty}{}
\makeatother
\usepackage{graphicx}
\usepackage{amsmath}
\usepackage{mathtools}
\usepackage{amssymb}
\usepackage{amsxtra}
\usepackage{booktabs}
\usepackage[noend]{algpseudocode}
\usepackage{bbding}
\usepackage{enumitem}
\usepackage{paralist}

\usepackage{algorithmicx,algorithm}
\usepackage{multirow}
\usepackage{colortbl}
\usepackage{color}
\usepackage{xcolor}
\usepackage{wrapfig}
\usepackage{changes}
\usepackage{amsmath}
\usepackage{enumerate}
\usepackage{xcolor,soul,framed} 
\usepackage{float}
\usepackage{hhline}
\usepackage{soul}
\usepackage{enumitem}

\usepackage[pagebackref,breaklinks,colorlinks]{hyperref}

\newcommand\sbtst{\bgroup\markoverwith{\textcolor{yellow}{\rule[0.5ex]{2pt}{0.5pt}}}\ULon}

\usepackage[capitalize]{cleveref}
\crefname{section}{Sec.}{Secs.}
\Crefname{section}{Section}{Sections}
\Crefname{table}{Table}{Tables}
\crefname{table}{Tab.}{Tabs.}

\def\cvprPaperID{7154} 
\def\confName{CVPR}
\def\confYear{2023}

\begin{document}

\title{Bi3D: Bi-domain Active Learning for Cross-domain 3D Object Detection}

\author{Jiakang Yuan$^{*,1}$, Bo Zhang$^{\dagger,2}$, Xiangchao Yan$^2$, Tao Chen$^{\dagger,1}$, Botian Shi$^2$, Yikang Li$^2$, Yu Qiao$^2$\\
$^1$School of Information Science and Technology, Fudan University \\ $^2$Shanghai AI Laboratory\\
{\tt\small {jkyuan22@m.fudan.edu.cn, \{yanxiangchao,shibotian,liyikang,qiaoyu\}@pjlab.org.cn}}
}


\maketitle

\newcommand\blfootnote[1]{%
\begingroup
\renewcommand\thefootnote{}\footnote{#1}%
\endgroup
}
 \blfootnote{{$^*$}This work was done when Jiakang Yuan was an intern at Shanghai AI Laboratory.} 
 \blfootnote{{$^\dagger$}Corresponding to: Tao Chen (eetchen@fudan.edu.cn), Bo Zhang (zhangbo@pjlab.org.cn)}

\begin{abstract}
Unsupervised Domain Adaptation (UDA) technique has been explored in 3D cross-domain tasks recently. Though preliminary progress has been made, the performance gap between the UDA-based 3D model and the supervised one trained with fully annotated target domain is still large. This motivates us to consider selecting partial-yet-important target data and labeling them at a minimum cost, to achieve a good trade-off between high performance and low annotation cost. To this end, we propose a Bi-domain active learning approach, namely Bi3D, to solve the cross-domain 3D object detection task. The Bi3D first develops a domainness-aware source sampling strategy, which identifies target-domain-like samples from the source domain to avoid the model being interfered by irrelevant source data. Then a diversity-based target sampling strategy is developed, which selects the most informative subset of target domain to improve the model adaptability to the target domain using as little annotation budget as possible. Experiments are conducted on typical cross-domain adaptation scenarios including cross-LiDAR-beam, cross-country, and cross-sensor, where Bi3D achieves a promising target-domain detection accuracy ($89.63\%$ on KITTI) compared with UDA-based work ($84.29\%$), even surpassing the detector trained on the full set of the labeled target domain ($88.98\%$). Our code is available at: \textit{\textcolor{teal}{\url{https://github.com/PJLab-ADG/3DTrans}}}.
\end{abstract}

\section{Introduction}
\label{sec:intro}

\begin{figure}[t]
\centering

\includegraphics[width=8.3cm]{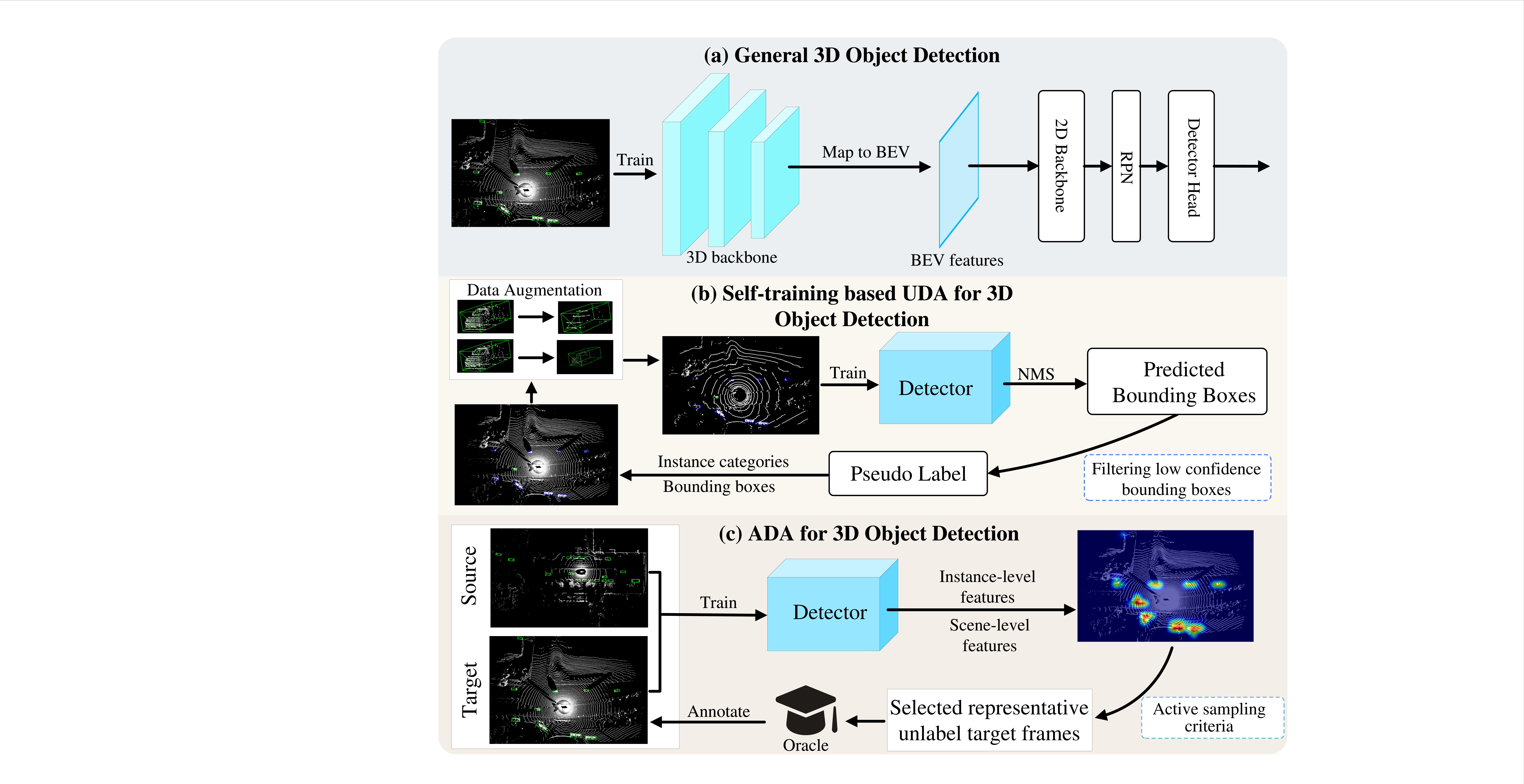}
\caption{Comparisons among (a) The general 3DOD pipeline, (b) Self-training based Unsupervised Domain Adaptation 3DOD pipeline, and (c) Active Domain Adaptation 3DOD pipeline that selects representative target data, and then annotates them by an oracle (human expert) for subsequent model refinement.}
\vspace{-0.2cm}
\label{fig:intro}
\end{figure}




LiDAR-based 3D Object Detection (3DOD)~\cite{lang2019pointpillars, shi2020pv, deng2021voxel, yan2018second, shi2020points} has advanced a lot recently. However, the generalization of a well-trained 3DOD model from a source point cloud dataset (domain) to another one, namely cross-domain 3DOD, is still under-explored. Such a task in fact is important in many real-world applications. For example, in the autonomous driving scenario, the target scene distribution frequently changes due to unforeseen differences in dynamically changing environments, making cross-domain 3DOD an urgent problem to be resolved.

Benefiting from the success of Unsupervised Domain Adaptation (UDA) technique in 2D cross-domain tasks~\cite{chen2018domain,inoue2018cross,zhang2021joint,tzeng2017adversarial,ganin2015unsupervised,long2016unsupervised,zou2018unsupervised}, several attempts are made to apply UDA for tackling 3D cross-domain tasks~\cite{luo2021unsupervised, saltori2020sf, zhang2021srdan, xu2021spg, yang2021st3d, wei2022lidar, qin2019pointdan}. 
ST3D~\cite{yang2021st3d} designs a self-training-based framework to adapt a pre-trained detector from the source domain to a new target domain. LiDAR distillation~\cite{wei2022lidar} exploits transferable knowledge learned from high-beam LiDAR data to the low one. Although these UDA 3D models have achieved significant performance gains for the cross-domain task, there is still a large performance gap between these UDA models and the supervised ones trained using a fully-annotated target domain. For example, ST3D~\cite{yang2021st3d} only achieves $72.94\%$ $\text{AP}_\text{3D}$ in nuScenes~\cite{caesar2020nuscenes}-to-KITTI~\cite{geiger2012we} cross-domain setting, yet the fully-supervised result using the same baseline detector can reach to $82.50\%$ $\text{AP}_\text{3D}$ on KITTI. 

To further reduce the detection performance gap between UDA-based 3D models and the fully-supervised ones, an initial attempt is to leverage Active Domain Adaptation (ADA) technique~\cite{xie2022active, xie2022learning, su2020active, fu2021transferable, prabhu2021active}, whose goal is to select a subset quota of all unlabeled samples from the target domain to perform the manual annotation for model training. Actually, the ADA task has been explored in 2D vision fields such as AADA~\cite{su2020active}, TQS~\cite{fu2021transferable}, and CLUE~\cite{prabhu2021active}, but its research on 3D point cloud data still remains blank. In order to verify the versatility of 2D image-based ADA methods towards 3D point cloud, we conduct extensive attempts by integrating the recently proposed ADA methods, \eg, TQS~\cite{fu2021transferable} and CLUE~\cite{prabhu2021active}, into several typical 3D baseline detectors, \eg, PV-RCNN~\cite{shi2020pv} and Voxel R-CNN~\cite{deng2021voxel}. Results show that these 2D ADA methods cannot obtain satisfactory detection accuracy under the 3D scene's domain discrepancies. For example, PV-RCNN coupled with TQS only achieves $75.40\%$ $\text{AP}_\text{3D}$, which largely falls behind the fully-supervised result $82.50\%$ $\text{AP}_\text{3D}$. 

As a result, directly selecting a subset of given 3D frames using 2D ADA methods to tackle 3D scene's domain discrepancies is challenging, which can be attributed to the following reasons. (1) The \textbf{sparsity} of the 3D point clouds leads to huge inter-domain discrepancies that harm the discriminability of domain-related features. (2) The \textbf{intra-domain feature variations} are widespread within the source domain, which enlarges the differentiation between the selected target domain samples and the entire source domain samples, bringing negative transfer to the model adaptation on the target domain.

To this end, we propose a Bi-domain active learning (Bi3D) framework to conduct the active learning for the 3D point clouds. To tackle the problem of \textbf{sparsity}, we design a foreground region-aware discriminator, which exploits an RPN-based attention enhancement to derive a foreground-related domainness metric, that can be regarded as an important proxy for active sampling strategy. To address the problem of \textbf{intra-domain feature variations} within the source domain, we conceive a Bi-domain sampling approach, where Bi-domain means that data from both source and target domains are picked up for safe and robust model adaptation. Specifically, the Bi3D is composed of a domainness-aware source sampling strategy and a diversity-based target sampling strategy. The source sampling strategy aims to select target-domain-like samples from the source domain, by judging the corresponding domainness score of each given source sample. Then, the target sampling strategy is utilized to select diverse but representative data from the target domain by dynamically maintaining a similarity bank. Finally, we employ the sampled data from both domains to adapt the source pre-trained detector on a new target domain at a low annotation cost.

The main contributions can be summarized as follows:
\begin{compactitem}
    \item[1.]
    From a new perspective of chasing high performance at a low cost, we explore the possibilities of leveraging active learning to achieve effective 3D scene-level cross-domain object detection.
    
    \item[2.]
    A Bi-domain active sampling approach is proposed, consisting of a domainness-aware source sampling strategy and a diversity-based target sampling strategy to identify the most informative samples from both source and target domains, boosting the model's adaptation performance.
    \item[3.]
    Experiments show that Bi3D outperforms state-of-the-art UDA works with only $1\%$ target annotation budget for cross-domain 3DOD. Moreover, Bi3D achieves $89.63\%$ $\text{AP}_\text{BEV}$ in the nuScenes-to-KITTI scenario, surpassing the fully supervised result ($88.98\%$ $\text{AP}_\text{BEV}$) on the KITTI dataset.
\end{compactitem}





\section{Related Works}

\subsection{LiDAR-based General and UDA 3D Detection}
LiDAR-based 3D object detection~\cite{chen2017multi, deng2021voxel, lang2019pointpillars, qi2018frustum, shi2019pointrcnn, shi2020points, shi2020pv, yan2018second,yang2018pixor, yang2019std, zhou2018voxelnet} has attracted increasing attention in real applications such as autonomous driving and robotics. Grid-based methods~\cite{yan2018second, deng2021voxel} convert disordered point cloud data to regular grids and extract features by 2D/3D convolution. Inspired by PointNet~\cite{qi2017pointnet}, Point-based approaches~\cite{yang2019std, shi2019pointrcnn} use set abstraction to extract features and directly generate proposals from point cloud data. However, these general 3D detectors still face serious performance drops in cross-domain applications, \eg, from Waymo or nuScenes to KITTI adaptation scenarios. UDA 3D object detection tackles the cross-domain distribution shift issue by various unsupervised methods. ST3D~\cite{yang2021st3d} proposes to use self-training and curriculum data augmentation to generate pseudo labels on a target domain to mitigate the large domain gap. LiDAR Distillation~\cite{wei2022lidar} proposes a distillation-based method, focusing on the knowledge transfer from high-beam data to low-beam data. However, there is still a large detection accuracy gap between these UDA methods~\cite{yang2021st3d, wei2022lidar} and fully-supervised 3D detectors~\cite{shi2020points, shi2020pv, yan2018second}.

\subsection{Active Domain Adaptation}
Inspired by active learning methods~\cite{sener2018active, joshi2012scalable, dasgupta2008hierarchical, Schohn2000LessIM, seung1992query, wang2014new, wang2016cost,yuan2021multiple} which aim to achieve relatively high recognition accuracies only using a small portion of informative data, Active Domain Adaptation (ADA)~\cite{su2020active, fu2021transferable, prabhu2021active, ning2021multi,xie2022active} has emerged in 2D vision task, which selects the most informative target data for annotation and adapts the model to the target domain by training on the selected data. CLUE~\cite{prabhu2021active} proposes to use an uncertainty-weighted clustering strategy to select informative target data. TQS~\cite{fu2021transferable} utilizes a hierarchical sampling strategy that performs active learning from multi-grained criteria such as transferable committee, transferable uncertainty, and transferable domainness. 

Although the ADA technique has achieved great success in 2D image tasks, its exploration of 3D point cloud tasks is still insufficient. Furthermore, it is intractable to directly apply these 2D ADA methods to the 3D point cloud scenarios, since these 2D ADA works~\cite{su2020active, fu2021transferable, prabhu2021active, ning2021multi} are not intended to tackle the distribution difference of point clouds with various spatial and geometric structures. Besides, previous ADA methods focus more on how to select samples from the target domain, ignoring that the source domain may contain many diverse samples and not all of them are beneficial for model adaptation to the target domain. In contrast, our Bi3D provides a new angle of view for achieving cross-domain generalization: a Bi-domain active learning strategy, which samples informative frames from both source and target domains.


\begin{figure*}
\vspace{-6pt}
\centering
\includegraphics[width=16.3cm]{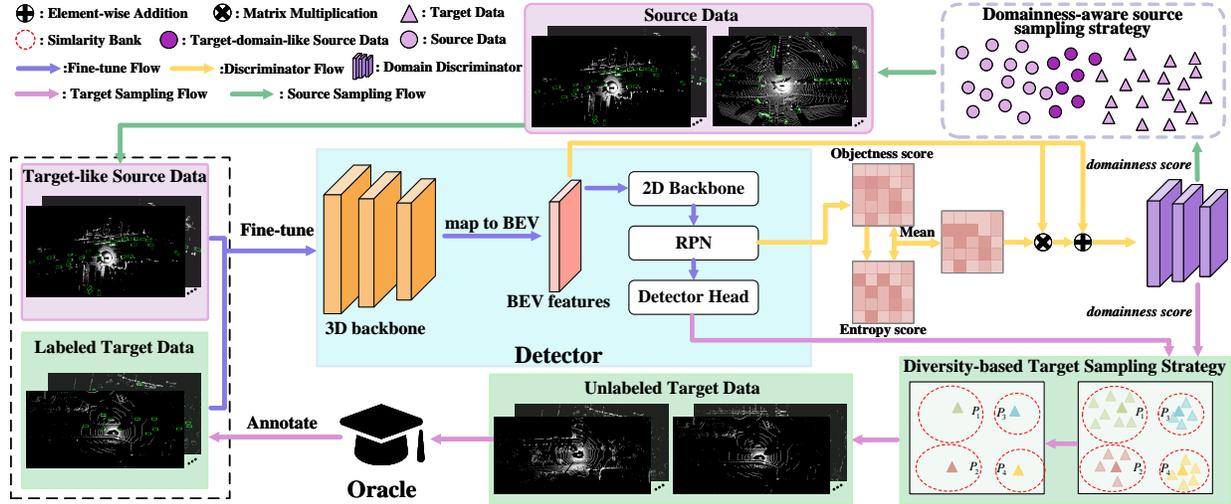}
\caption{The overview of the proposed Bi3D, which employs PV-RCNN as our baseline and consists of domainness-aware source sampling strategy and diversity-based target sampling strategy. The target-domain-like source data are first selected by the learned domainness score, and then the detector is fine-tuned on the selected source domain data. Next, diverse and representative target data are selected using a similarity bank, and then annotated by an oracle. Finally, the detector is fine-tuned on both the selected source and target data.}
\vspace{-0.3cm}
\label{fig1}
\end{figure*}

\vspace{-0.2cm}
\section{Method}
\label{sec:method}
\vspace{-0.1cm}
The overall Bi3D framework is shown in Fig. \ref{fig1}. To better illustrate the Bi3D principle, we first describe our problem definition and the selected baseline model. Next, we introduce the proposed Bi3D. Finally, we give the overall objectives and Bi-domain sampling and training strategies.
\vspace{-0.2cm}
\subsection{Preliminary}
\vspace{-0.1cm}
\noindent\textbf{Problem Definition.} Given a labeled source domain set {\small $D_s=\{(\mathbf{x}_i^s, y_i^s)\}_{i=1}^{n_s}$}, an unlabeled target domain set {\small $D_t=\{\mathbf{x}_j^t\}_{j=1}^{n_t}$}, and an annotation budget $B$, where $B\ll n_t$ and $n_t$ denotes the total amount of target domain data. Following the standard ADA setting, a labeled target dataset {\small $\Tilde{D_t}$} is constructed, which is initially empty and will be updated in $R$ rounds of the sampling process. In the $k$-th sampling round where {\small $k\leq R$}, a subset {\small $\Delta D_t^k$} is selected from {\small $D_t/\Tilde{D_t}$} and labeled by an oracle (\textbf{human expert)}. Then, {\small $\Tilde{D_t}$} will be updated as {\small $\Tilde{D_t}\leftarrow \Tilde{D_t}\cup \Delta D_t^k$}. After {\small $R$} rounds of sampling, the number of data in $\Tilde{D_t}$ reaches the upper limit of annotation budget $B$, \textit{i.e.}, {\small $|\Tilde{D_t}|=B$}. Note that different from previous ADA methods, in this work, we further construct a source subset {\small $\Tilde{D_s}$} sampled from the original source domain {\small $D_s$}. The goal of the proposed Bi3D is to select both target-domain-like data from {\small $D_s$} and the most informative data from $D_t$, to constitute {\small $\Tilde{D_s}$} and {\small $\Tilde{D_t}$}, and make the 3D detector better adapt to target domain by jointly training on a mixture set from {\small $\Tilde{D_s}$} and {\small $\Tilde{D_t}$}.

\noindent\textbf{Baseline Introduction.} Following previous cross-domain studies~\cite{yang2021st3d, wei2022lidar} for 3DOD, we use PV-RCNN~\cite{shi2020pv} as our baseline model. PV-RCNN is a typical two-stage 3D detection framework that takes advantage of both the point-based network and 3D voxel-based CNN. The overall loss function of PV-RCNN can be written as follows:
\vspace{-0.10cm}
\begin{equation}
    L_{det}=L_{rpn}+L_{rcnn}+L_{seg}
    \label{eq:pvrcnn_loss}
\vspace{-0.10cm}
\end{equation}

\noindent where {\small $L_{rpn}$} denotes the loss of Region Proposal Network (RPN), {\small$L_{rcnn}$} represents the proposal refinement loss and {\small $L_{seg}$} is the keypoint segmentation loss.

\subsection{Bi3D: Bi-domain Active Learning for 3D Object Detection}
\label{subsec:method}

To effectively measure the domainness of source and target samples, we first design a foreground region-aware discriminator. Then, based on the domain discriminator, we propose a Bi-domain sampling strategy to adapt a pre-trained 3D detector from its source domain to a new target domain.

\noindent\textbf{Foreground Region-aware Discriminator.} Considering that instance-level features lose the contextual relationship between the instance and its original scene, and meanwhile, a large number of negative anchors will greatly hinder domain discriminator learning, we thus generate scene-level  representations by extracting from Bird-Eye-View (BEV) features. However, the BEV features extracted using 3D convolution are very sparse due to the sparse distribution of point cloud data, causing the traditional discriminator difficult to localize and learn on informative foreground regions, thus resulting in a biased domain representation learning.


To address this issue, we design a foreground region-aware discriminator, aiming at measuring the frame-level domainness score for both the source and target data by enhancing foreground-region features in the scene. Specifically, let {\small $\mathbf{x}_i^d$} denote the input point cloud data, where {\small $d\in[s,t]$} means that the sample {\small $\mathbf{x}$} is from source domain $s$ or target domain $t$. Next, the 3D feature volumes are first encoded by the 3D backbone {\small $F_{3D}$} and then converted into 2D BEV features {\small $f_{bev}\in\mathbb{R}^{C\times H \times W}$}, where {\small $C$} denotes the channel number, {\small $H$} and {\small$W$} are the height and width of the feature, respectively. 

To make the domain discriminator pay more attention to foreground regions, we first obtain the objectness score {\small $\mathit{S}_{obj} \in \mathbb{R}^{C\mathrm{'}\times H \times W}$} by the RPN operation, where {\small$C\mathrm{'}$} indicates the number of anchors per location. The objectness score represents the probability that a default anchor belongs to a foreground object. 
In order to quantitatively evaluate the prediction uncertainty of the detector for the current scene, inspired by previous methods~\cite{holub2008entropy, fu2021transferable,qiu2016maximum}  using entropy to measure uncertainty, we calculate the entropy score {\small $\mathit{S}_{ent} \in \mathbb{R}^{C{'}\times H \times W}$} with the following formula:
\begin{equation}
\label{entropy}
    \mathit{S}_{ent}=-\mathit{S}_{obj}\log\mathit{S}_{obj} - (1-\mathit{S}_{obj})\log (1-\mathit{S}_{obj}),
\end{equation}
where {\small $\mathit{S}_{ent}$} denotes the uncertainty of a spatial location being classified as an instance object.
Based on Eq.~\ref{entropy}, a scene-level attention map can be obtained by combining {\small $\mathit{S}_{obj}$} and {\small $\mathit{S}_{ent}$}, which can make the model pay more attention to foreground features. Thus, the foreground region-aware features can be calculated as follows:
\begin{equation}
    \hat{f}_{bev}=(1+(\mathit{\hat{S}}_{obj}+\mathit{\hat{S}}_{ent})/2) \ f_{bev},
\end{equation}
where {\small $\hat{f}_{bev}$} represents the foreground region-aware BEV features, and {\small $\mathit{\hat{S}}_{obj}$} and {\small $\mathit{\hat{S}}_{ent}$} are the maximum value of {\small $\mathit{S}_{obj}$} and {\small $\mathit{S}_{ent}$} along the channel dimension, respectively.

Based on the foreground region-aware feature maps {\small $\hat{f}_{bev}$}, a domain discriminator with a convolution block is utilized to classify whether the data is from the source domain or target domain. For a detailed structure of the discriminator, please refer to our supplementary material. The loss function of the domain discriminator can be written as follows:
\begin{equation}
\label{dom}
\begin{aligned}
    L_{dom}=&-\mathbb{E}_{x^s\sim D_s}[\log(1-H({\hat{f}_{bev}}^s)]\\ &-\mathbb{E}_{x^t\sim D_t}[\log(H(\hat{f}_{bev}^t)],
\end{aligned}
\end{equation}
where {\small $L_{dom}$} is the domainness loss, and {\small $H$} denotes the domain discriminator, where we label the source domain and the target domain as '0' and '1', respectively.

\noindent\textbf{Domainness-aware Source Sampling Strategy.}
Previous DA works mainly focus on how to fully exploit representative data from the target domain, which actually ignores that there are a certain number of source samples interfering with the target domain representation learning. Thus, we propose a simple but effective domainness-aware source sampling strategy, aiming at selecting target-domain-like samples from the source domain to initially strengthen the model adaptability. 
In particular, we first calculate scene-level domainness score $s_i^s$ of all source data using the aforementioned domain discriminator, where $s_i^s={\small H(\hat{f}_{bev}^s)}$ and $s_i^s$ can be regarded as a similarity metric between source data and target data. $s_i^s$ with a relatively high value indicates that the $i$-th frame data from the source domain complies with the data distribution of the target domain. To select the source data with a high domainness score, we \underline{\textbf{simply sort $s_i^s$ in descending order}}, thus {\small $\Tilde{D}_s$} can be built by sampling the sorted data with a proportion or threshold. Please refer to our supplementary material for the study of the number of selected source data. Note that there is a smaller domain gap between {\small $\Tilde{D}_s$} and {\small $D_t$} and therefore by fine-tuning the detector on {\small $\Tilde{D}_s$}, the performance of the model on the target domain will be improved. As a result, the detector can extract more accurate instance-level features, benefiting to select more informative target data.

\noindent\textbf{Diversity-based Target Sampling Strategy.} To make the detector better adapt to the target domain, we first fine-tune the detector on {\small $\Tilde{D}_s$}, and select representative data from the target domain. However, since the adjacent frames in scenes like autonomous driving are usually similar, traditional active learning methods (\eg, Query-by-Committee~\cite{Schohn2000LessIM}, Query-by-Uncertainty~\cite{wang2014new}) encounter a major challenge that they often select the samples with a small between-class difference, causing redundant frame annotation operations. Thus, we design a diversity-based target sampling strategy to select diverse-and-representative target domain data.

Given ROI features {\small $\mathbf{I}_j=[I_j^1,I_j^2,...,I_j^k]$} from the $j$-th target frame, the corresponding confidence scores {\small $\mathbf{d}_j=[d_j^1,d_j^2,...,d_j^k]$} can be easily obtained by the baseline detector with the standard post-processing process, \textit{i.e.,} Non-Maximum Suppression (NMS). We first use confidence scores to re-weight all ROI instance-level features to obtain more accurate instance descriptions {\small $\hat{I}_j$} in the current frame $\mathbf{x}_j^t$, where {\small $\hat{I}_j=\mathbf{I}_j^\mathrm{T}\mathbf{d}_j$}, and the domainness score of target domain $s_j^t$ can be calculated by the designed domain discriminator {\small $H$} described above. As summarized in Algorithm \ref{alg:alg_1}, the basic idea of the diversity-based target sampling strategy is to maintain a similarity bank, where all unlabeled target data are clustered based on pairwise similarity of re-weighted ROI features to ensure the diversity of selected target data. In particular, we use cosine distance to measure the similarity $\alpha$ and dynamically update the prototypes of candidate ROI features using the following formula:
\begin{equation}
    \label{eq:center}
    \hat{c}(P_m, P_n)=\frac{num(P_m)\times c_m+num(P_n)\times c_n}{num(P_m)+num(P_n)},
\end{equation}
where $c_m$, $c_n$ are the $m$-th and $n$-th prototypes assigned according to the preset budget, meaning that each budget is represented by one prototype. {\small $P_m$} and {\small $P_n$} denote the similarity bank of the above $m$-th and $n$-th budget-wise prototypes, which are used to buffer unlabeled frames, and {\small $num(\cdot)$} denotes the number of unlabeled frames in the buffer. After Algorithm~\ref{alg:alg_1} is finished, to sample more diverse and representative frames from the target domain, we select one unlabeled frame $x_j^t$ with the top-$1$ domainness score $s_j^t$ from each updated bank $\mathbf{P}$, to form the full set of all data for manual annotation.

\begin{algorithm}[t]
\caption{Diversity-based Target Sampling Strategy}
\label{alg:alg_1}
\hspace*{0.02in} {\bf Input:}
 The $j$-th unlabeled frame $x_j^t$, where $x_j^t \in D_t/\Tilde{D}_t$,\\
\hspace*{0.44in} and the $k$-th round of sampling budget $b_k$\\
\hspace*{0.02in} {\bf Output:} 
The selected target set $\Delta D_t^k$
\begin{algorithmic}[1]
\State Calculate the re-weighted ROI features ${\hat{I}_j}$ with the obtained domainness score $s_j^t$ from the unlabeled frame $x_j^t$, by $s_j^t={\small H(\hat{f}_{bev}^t)}$.
\State Initialize the similarity bank $\mathbf{P}\coloneqq\emptyset$ and budget prototypes $\mathbf{c}\coloneqq\emptyset$
\For{$x_j^t$ in $D_t/\Tilde{D}_t$} 
    \If{$|\Delta D_t^k|<b_k$}
        \State Update $\mathbf{P}\coloneqq \mathbf{P}\cup x_j^t$, $\mathbf{c}\coloneqq \mathbf{c}\cup \hat{I}_j$
    \Else
        \State Calculate the similarity $\alpha_{\hat{I}_j,\mathbf{c}}$ between $\hat{I}_j$ and $\mathbf{c}$
        \State Calculate the similarity $\alpha_{\mathbf{c}}$ of prototypes in $\mathbf{c}$

        \If{$\max(\alpha_{\hat{I}_j,\mathbf{c}})<\min(\alpha_{\mathbf{c}})$}
            \State Merge the most similar banks $P_m$ and $P_n$ \State and the corresponding prototypes $c_m$ and $c_n$ \State using Eq. \ref{eq:center}
            \State Update $\mathbf{P}\coloneqq \mathbf{P}\cup x_j^t$, $\mathbf{c}\coloneqq \mathbf{c}\cup \hat{I}_j$
        \Else
            \State Merge $\hat{I}_j$ into $P_m$, where the corresponding \State prototype $c_m$ is most similar to $\hat{I}_j$ 
        \EndIf
    \EndIf
\EndFor
\State Select data in each bank by $s_j^t$ and fill the $\Delta D_t^k$
\State \Return Selected target subset $\Delta D_t^k$
\end{algorithmic}
\end{algorithm}

\subsection{Overall Objective and Bi-domain Sampling and Training Strategy}
\noindent\textbf{Overall Objectives.} The overall objective can be formulated as follows:
\begin{equation}
    L_{det}=\mathbb{E}_{x\sim \Tilde{D}_s\cup \Tilde{D}_t}[L_{rpn}+L_{rcnn}+L_{seg}],
\end{equation}
where the definition of $L_{rpn}$, $L_{rcnn}$, $L_{seg}$ follows Eq. \ref{eq:pvrcnn_loss}.

\noindent\textbf{Bi-domain Sampling and Training Strategy.} To adapt the detector from the source domain to the target domain, our method includes four steps. \textit{1) Pre-training on source domain:} The detector is firstly pre-trained on {\small $D_s$} using Eq.~\ref{eq:pvrcnn_loss} to ensure that the detector can learn sufficient knowledge for model transfer. \textit{2) Training the domain discriminator:} We freeze the parameters from the baseline detector while training the designed domain discriminator using $L_{dom}$ in Eq.~\ref{dom}. \textit{3) Active sampling source domain:} In this step, we select target-domain-like source data and fine-tune the detector on {\small $\Tilde{D}_s$} to reduce the domain gap. \textit{4) Active sampling target domain:} Based on the selected source data and the fine-tuned detector, we further sample the most informative target data and re-train the detector on both {\small $\Tilde{D}_s$} and {\small $\Tilde{D}_t$}.

\begin{table*}[htbp]
    \centering
    \begin{small}
        \begin{tabular}{c|c|>{\columncolor{gray!10}}c|c|>{\columncolor{gray!10}}c|c}
            \bottomrule[1pt]
            \multirow{2}{*}{Task} & \multirow{2}{*}{Method}  & \multicolumn{2}{c|}{PV-RCNN} & \multicolumn{2}{c}{Voxel R-CNN} \\
            \hhline{~~|-|-|-|-}
            &  & $\text{AP}_{\text{BEV}}$ / $\text{AP}_{\text{3D}}$ & Closed Gap & $\text{AP}_{\text{BEV}}$ / $\text{AP}_{\text{3D}}$ & Closed Gap \\
            \hline
            \multirow{10}{*}{Waymo$\rightarrow$KITTI} & Source Only  & 61.18 / 22.01 & - & 64.87 / 19.90 & - \\
            & ST3D~\cite{yang2021st3d} & 84.10 / 64.78 & +82.45\% / +70.71\% &  65.67 / 20.14 & +03.26\% / +00.38\%  \\
            & Ours ($1\%$) & \textbf{85.13} / \textbf{71.36} & +86.15\% / +81.58\% & \textbf{86.35} / \textbf{72.70} & +87.42\% / +83.36\% \\
            \hhline{~|-|-|-|-|-}
            & SN \cite{wang2020train} & 79.78 / 63.60 & +66.91\% / +68.76\% & 71.65 / 61.63 & +27.55\% / +65.88\% \\
            & ST3D (w/ SN)~\cite{yang2021st3d} & 86.65 / 76.86 & +91.62\% / +90.68\% & 80.23 / 68.98 & +62.52\% / +77.49\% \\
            & CLUE (w/ SN, $1\%$)~\cite{prabhu2021active} & 82.13 / 73.14 & +75.36\% / +84.53\%  & 81.93 / 70.89 & +69.43\% / +80.50\% \\
            & TQS (w/ SN, $1\%$)~\cite{fu2021transferable} & 82.00 / 72.04 & +74.89\% / +82.77\% & 78.26 / 67.11 & +54.50\% / +74.53\% \\
            & Ours (w/ SN, $1\%$) & \textbf{87.12} / \textbf{78.03} & +93.31\% / +92.61\% & \textbf{88.09} / \textbf{79.14} & +94.51\% / +93.53\% \\
            & Ours (w/ SN, $5\%$) & \textbf{89.53} / \textbf{81.32} & +102.64\% / +97.39\% & \textbf{90.18} / \textbf{81.34} & +103.01\% / +97.00\% \\

            \hhline{~|-|-|-|-|-}
            & Oracle &  88.98 / 82.50 & - & 89.44 / 83.24 & - \\
            \toprule[1pt]
            \bottomrule[1pt]
            \multirow{10}{*}{Waymo$\rightarrow$Lyft} & Source Only  & 75.49 / 58.53 &  &  70.52 / 53.48 & - \\
             & ST3D~\cite{yang2021st3d}  & 77.68 / 60.53 & +19.96\% / +15.20\% &  72.27 / 54.94 & +15.97\% / +21.22\% \\
            & Ours ($1\%$) & \textbf{79.06} / \textbf{63.70} & +32.54\% / +39.29\% & \textbf{78.39} / \textbf{64.50} & +71.81\% / +160.17\%  \\
            \hhline{~|-|-|-|-|-}
            & SN \cite{wang2020train}  & 72.82 / 56.64 & {\large-}24.34\% / {\large-}14.36\% & 68.77 / 52.67 & {\large-}15.97\% / {\large-}11.77\% \\
            & ST3D (w/ SN)~\cite{yang2021st3d} & 74.95 / 58.54 & {\large-}04.92\% / +00.08\% &  69.91 / 54.23 & {\large-}05.57\% / +10.90\% \\
            & CLUE (w/ SN, $1\%$)~\cite{prabhu2021active} & 75.23 / 62.17 & {\large-}02.37\% / +27.66\% & 75.61 / 59.34 & +46.44\% / +85.17\% \\
            & TQS (w/ SN, $1\%$)~\cite{fu2021transferable} & 70.87 / 55.25 & {\large-}42.11\% / {\large-}24.92\% & 71.11 / 56.28 & +05.38\% / +40.70\% \\
            & Ours (w/ SN, $1\%$) & \textbf{79.07} / \textbf{63.74} & +32.63\% / +39.59\% & \textbf{77.00} / \textbf{61.23} & +59.12\% / +112.65\% \\
            & Ours (w/ SN, $5\%$) & \textbf{80.12} / \textbf{65.54} & +42.21\% / +53.27\% & \textbf{79.15} / \textbf{65.26} & +78.74\% / +171.22\% \\
            \hhline{~|-|-|-|-|-}
            & Oracle &  86.46 / 71.69 & - &  81.48 / 60.36 & - \\
            \toprule[1pt]
            \bottomrule[1pt]
            \multirow{11}{*}{Waymo$\rightarrow$nuScenes} & Source Only  & 34.50 / 21.47 & - & 32.58 / 16.53 & - \\
            & ST3D~\cite{yang2021st3d} & 36.42 / 22.99 & +10.32\% / +08.89\% & 34.68 / 17.17 & +12.40\% / +03.33\% \\
            & LiDAR Distill~\cite{wei2022lidar} & 43.31 / 25.63 & +47.34\% / +24.34\% & - & - \\
            & Ours ($1\%$) & \textbf{45.52} / \textbf{30.75} & +59.22\% / +54.30\% & \textbf{44.86} / \textbf{29.52} & +72.45\% / +67.63\% \\
            \hhline{~|-|-|-|-|-|}
            & SN \cite{wang2020train} & 34.22 / 22.29 & {\large-}01.50\% / +04.80\% &  29.43 / 19.21 & {\large-}18.60\% / +13.95\% \\
            & ST3D (w/ SN)  \cite{yang2021st3d} & 36.62 / 23.67 & +11.39\% / +12.87\% & 32.77 / 22.21 & +01.12\% / +29.57\% \\
            & CLUE (w/ SN, $1\%$) \cite{prabhu2021active}  & 38.18 / 26.96 & +19.77\% / +32.12\%  & 37.27 / 25.12 & +39.49\% / +44.72\% \\
            & TQS (w/ SN, $1\%$)~\cite{fu2021transferable} & 35.47 / 25.00 & +05.01\% / +20.66\% & 36.38 / 24.18 & +22.43\% / +39.82\% \\
            & Ours (w/ SN, $1\%$) & \textbf{45.00} / \textbf{30.81} & +56.42\% / +54.65\% & \textbf{45.29} / \textbf{29.70} & +75.03\% / +68.56\% \\
            & Ours (w/ SN, $5\%$) & \textbf{48.03} / \textbf{32.02} & +72.70\% / +61.73\%  & \textbf{47.02} / \textbf{31.23} & +85.24\% / +76.52\% \\
            \hhline{~|-|-|-|-|-|}
            & Oracle &  53.11 / 38.56 & - &  49.52 / 35.74 & - \\
            \toprule[1pt]
            \bottomrule[1pt]
            \multirow{10}{*}{nuScenes$\rightarrow$KITTI} & Source Only & 68.15 / 37.17  & - & 67.27 / 30.54 & - \\
            & ST3D \cite{yang2021st3d} & 78.36 / 70.85 & +49.02\% / +74.30\% & 74.16 / 35.55 & +31.08\% / +09.51\% \\
            & Ours ($1\%$) & \textbf{84.91} / \textbf{71.56} & +80.64\% / +75.87\% & \textbf{86.10} / \textbf{72.75} & +84.93\% / +80.08\% \\
            \hhline{~|-|-|-|-|-}
            & SN \cite{wang2020train} & 60.48 / 49.47 & {\large-}36.82\% / +27.13\%  & 44.00 / 25.20 & {\large-}104.96\% / {\large-}10.13\% \\
            & ST3D (w/ SN) \cite{yang2021st3d} &  84.29 / 72.94 & +77.48\% / +78.91\% & 52.44 / 20.99 & {\large-}66.89\% / {\large-}18.12\% \\
            & CLUE (w/ SN, $1\%$) \cite{prabhu2021active} & 74.77 / 64.43 & +37.18\% / +60.14\% & 79.12 / 68.02 & +53.45\% / +71.12\% \\
            & TQS (w/ SN, $1\%$)~\cite{fu2021transferable} & 84.66 / 75.40 & +79.26\% / +84.34\% & 77.98 / 66.02 & +48.31\% / +67.32\% \\
            & Ours (w/ SN, $1\%$) & \textbf{87.00} / \textbf{77.55} & +90.49\% / +89.08\% & \textbf{87.33} / \textbf{77.24} & +90.48\% / +88.61\% \\
            & Ours (w/ SN, $5\%$) & \textbf{89.63} / \textbf{81.02} & +103.12\% / +96.73\% & \textbf{88.15} / \textbf{79.06} & +94.18\% / +92.07\% \\

            \hhline{~|-|-|-|-|-}
            & Oracle & 88.98 / 82.50 & - & 89.44 / 83.24 & - \\
            \toprule[0.8pt]
        \end{tabular}
    \end{small}
    \caption{Results on different adaptation scenarios under $1\%$ and $5\%$ annotation budget. Following ~\cite{yang2021st3d, wei2022lidar}, we report $\text{AP}_{\text{BEV}}$ and $\text{AP}_{\text{3D}}$ over 40 positions' recall for the car category at IoU = 0.7. \textbf{Source Only} denotes that the pre-trained detector is directly evaluated on the target domain, and \textbf{Oracle} represents the detection results obtained using the fully-annotated target domain. Closed Gap denotes the performance gap closed by various methods along Source Only and Oracle results. The best adaptation results are marked in \textbf{bold}.}
    \vspace{-0.3cm}
    \label{tab:SOTAcomparison}
\end{table*}

\section{Experiments}
\subsection{Experimental Setup}
\noindent\textbf{Datasets.} We conduct experiments on four popular autonomous driving datasets: KITTI~\cite{geiger2012we}, Waymo~\cite{sun2020scalability}, nuScenes~\cite{caesar2020nuscenes} and Lyft~\cite{lyft2019}. We consider four cross-domain settings including cross-LiDAR-beam (\textit{i.e.} Waymo-to-nuScenes, nuScenes-to-KITTI), cross-country (\textit{i.e.} Waymo-to-KITTI), and cross-sensor scenarios (\textit{i.e.} Waymo-to-Lyft). Following previous domain adaptation works~\cite{wei2022lidar,yang2021st3d}, we use the KITTI evaluation metric to perform all experiments on Car (Vehicle in Waymo) category.

\noindent\textbf{Implementation Details.} We evaluate the proposed Bi3D on two widely-used detectors: PV-RCNN~\cite{shi2020pv} and Voxel R-CNN~\cite{deng2021voxel}. Following~\cite{yang2021st3d,wei2022lidar}, we only use the coordinate encoding {\small $(x,y,z)$} of raw point cloud as the detector input, and set the voxel size of both PV-RCNN and Voxel R-CNN to {\small $(0.1m,0.1m,0.15m)$} on all datasets. \textbf{In the stage of active sampling source domain}, we first select the target-domain-like data from the source domain in the initial training epoch and fine-tune the detector for the following $15$ epochs. \textbf{In the stage of active sampling target domain}, we mainly consider the situation that the annotation budget $B$ is equal to $1\%$ and $5\%$, respectively, which follows the standard experimental setting in the ADA task. Our method is implemented using OpenPCDet~\cite{openpcdet2020}.


\subsection{Comparison Baselines}
To verify the effectiveness of the proposed Bi3D, we design several baseline methods including both active learning and active domain adaptation based methods.

\noindent\textbf{1) Random:} We randomly select the target domain data for performing the manual annotation.

\noindent\textbf{2) Entropy}~\cite{wang2014new}\textbf{:} By measuring the entropy of samples from the target domain, we select the samples with relatively high entropy scores, which can represent the sample-level uncertainty predicted by a detector.

\noindent\textbf{3) Committee}~\cite{Schohn2000LessIM}\textbf{:} By using multiple classifiers to predict the categories of target samples, the samples with inconsistent prediction scores along with all classifiers are selected.

\noindent\textbf{4) CLUE}~\cite{prabhu2021active}\textbf{:} CLUE is a representative work under ADA setting, which proposes Clustering Uncertainty-weighted Embeddings in order to select informative-and-diverse target data by means of a re-weighted uncertainty clustering.

\noindent\textbf{5) TQS}~\cite{fu2021transferable}\textbf{:} TQS is a prior work to explore the transferable criteria which are specially designed to mitigate the domain gap. TQS picks up data by combining a series of transferable sampling strategies (such as committee, uncertainty, and domainness) to reduce the sampling uncertainty.

\vspace{-0.1cm}
\subsection{Main Results}
\vspace{-0.1cm}
\noindent\textbf{Comparison with 2D ADA works.} To verify the effectiveness of our Bi3D and ensure the fairness of experiments, we first compare our method with two widely used 2D ADA methods (\textit{i.e.} CLUE and TQS) under the same cross-domain setting. As shown in Table \ref{tab:SOTAcomparison}, it can be seen that compared with 2D ADA methods, our Bi3D achieves better results by a large margin on all cross-domain scenarios, demonstrating the method's scalability for 3D point cloud detection tasks. Meanwhile, we can observe that 2D ADA methods cannot achieve satisfactory results, and even fall behind UDA methods (\textit{i.e.} Waymo$\rightarrow$KITTI on PV-RCNN). A detailed analysis is described in Section \ref{subsec:analysis}.

\noindent\textbf{Comparison with 3D UDA works.} We deeply review the cross-domain 3D object detection works~\cite{yang2021st3d, wei2022lidar}, and find that previous works mainly focus on the study of UDA 3D detection. To show the effectiveness of active learning, we compare our Bi3D with these cross-domain 3D detection works. For example, ST3D~\cite{yang2021st3d} uses self-training to iteratively improve the performance on the target domain, LiDAR Distillation~\cite{wei2022lidar} generates the low-beam pseudo point cloud and distills the knowledge from high-beam data, which achieves state-of-the-art results on high-to-low beam adaptation scenario. It can be seen from Table~\ref{tab:SOTAcomparison} that, the Bi3D greatly reduces the performance gap between different domains, surpassing all state-of-the-art UDA 3DOD methods. Note that the Bi3D largely improves the performance on the difficult Waymo$\rightarrow$nuScenes setting ($\text{AP}_{\text{BEV}}$: $36.42\%$ $\rightarrow$ $45.52\%$ compared to ST3D, and $43.31\%$ $\rightarrow$ $45.52\%$ compared to LiDAR Distillation, $\text{AP}_{\text{3D}}$: $22.99\%$ $\rightarrow$ $30.75\%$ compared to ST3D, and $25.63\%$ $\rightarrow$ $30.75\%$ compared to LiDAR Distillation). Besides, our experiments are conducted under $1\%$ target annotation budget, demonstrating that Bi3D can largely improve the cross-domain detection performance at a low annotation cost.

\noindent\textbf{Comparison with 3D weakly-supervised DA works.} SN~\cite{wang2020train} is a typical weakly-supervised DA method, which uses statistic-level normalization to reduce the domain difference caused by source-to-target object size variances. We conduct experiments combining our method with SN.
The results are reported in Table \ref{tab:SOTAcomparison}, which show that our method outperforms all methods with SN operation. We find that the result can be further improved especially on cross-country adaptation setting (\textit{i.e.} $71.56\%$ $\rightarrow$ $77.55\%$ on nuScenes$\rightarrow$KITTI and $71.36\%$ $\rightarrow$ $78.03\%$ on Waymo$\rightarrow$KITTI). This is mainly because SN can reduce the domain shift caused by object size variations and is beneficial to pick up more target-domain-like source data.

\subsection{Insightful Analyses}
\label{subsec:analysis}

\begin{figure}
  
  \begin{subfigure}{0.49\linewidth}
    \includegraphics[width=4.1cm]{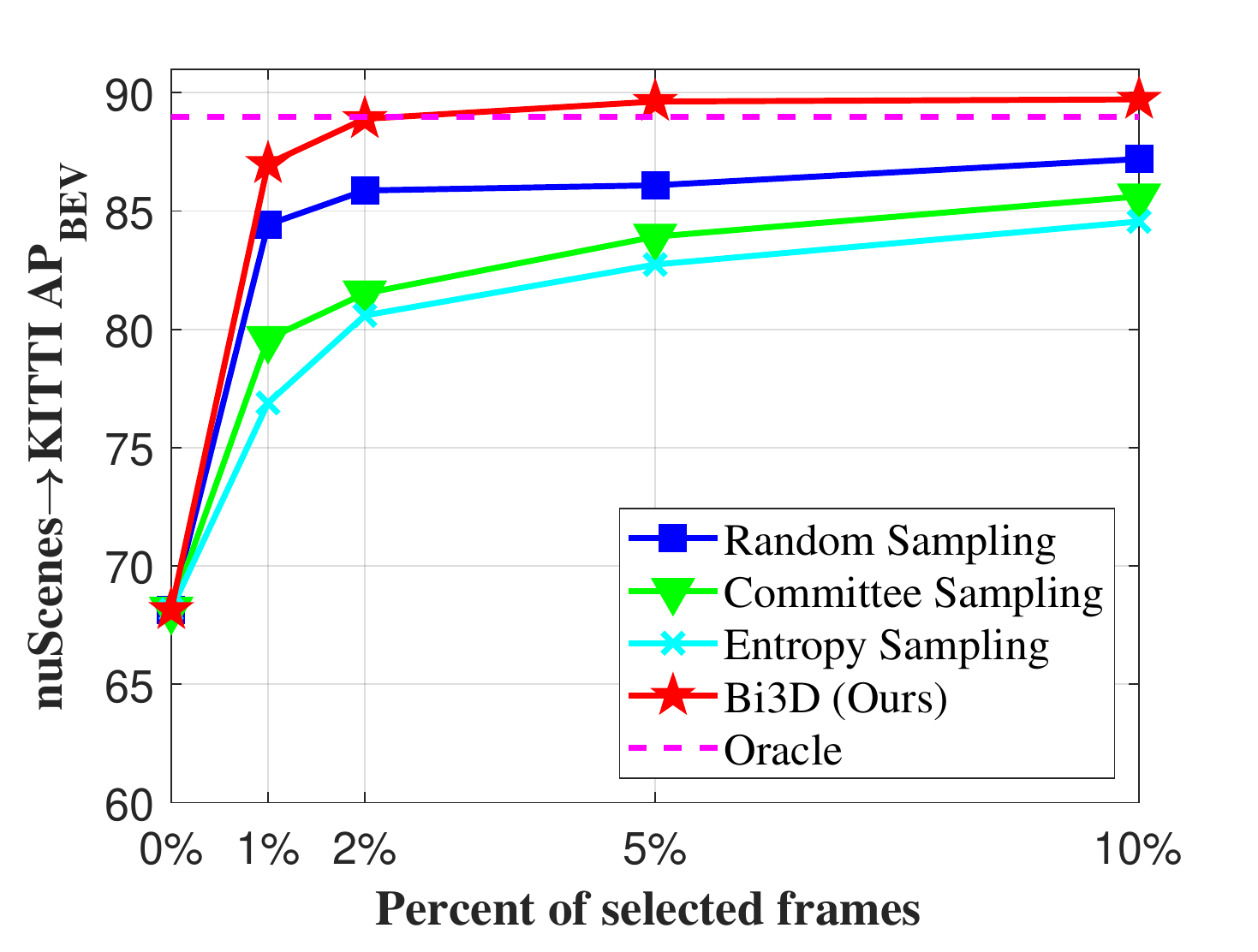}
    \caption{nuScenes$\rightarrow$KITTI}
    \label{fig:subfig_a}
  \end{subfigure}
  \begin{subfigure}{0.49\linewidth}
    \includegraphics[width=4.1cm]{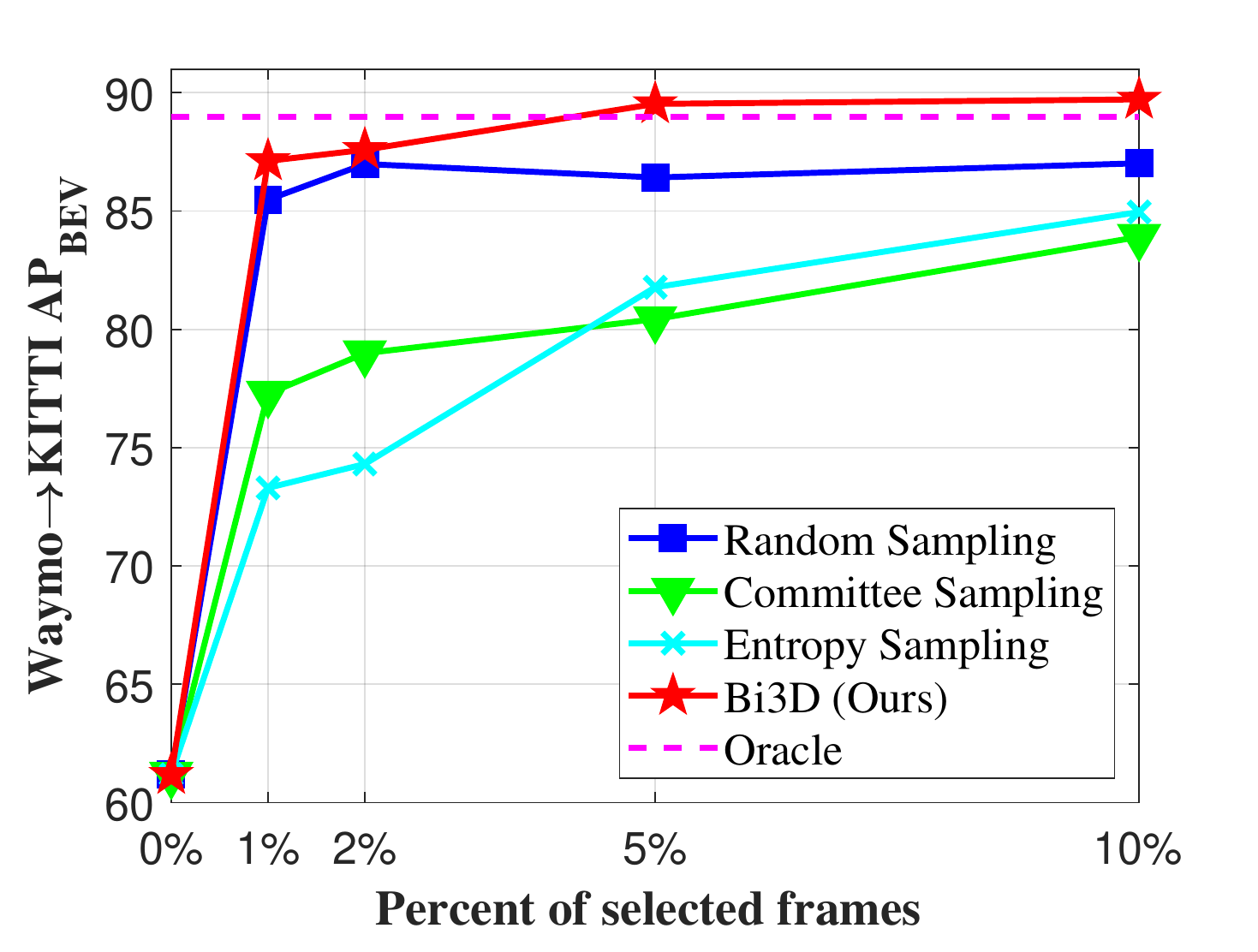}
    \caption{Waymo$\rightarrow$KITTI}
    \label{fig:subfig_b}
  \end{subfigure}
  \caption{Results of various target annotation budgets.}
  \label{fig:budget}
\vspace{-0.35cm}
\end{figure}

\noindent\textbf{Results of Changing Target Domain Annotation Budget.} In this part, we compare our Bi3D with several typical active learning methods (\textit{i.e.} query-by-committee~\cite{Schohn2000LessIM} and query-by-uncertainty~\cite{wang2014new}), and their results demonstrate that Bi3D can consistently outperform all these methods. We also conduct experiments on nuScenes$\rightarrow$KITTI and Waymo$\rightarrow$KITTI by changing the annotation budget. As illustrated in Fig. \ref{fig:budget}, we plot the trend of $\text{AP}_\text{BEV}$ at different manual annotation budgets.
It can be seen that our Bi3D achieves a promising detection accuracy gain, even outperforming many active learning methods. Besides, with the increase of the number of manually annotated target frames, the model detection accuracy is constantly improved. Furthermore, when the manually labeled target data reaches $5\%$ of the total number of unlabeled frames, Bi3D can greatly improve the cross-domain detection accuracy of the baseline detector, even surpassing the fully-supervised results with $100\%$ labeled target data.

As described above, we found that when 2D ADA works (as shown in~Table~\ref{tab:SOTAcomparison}) and 2D active learning works (as illustrated in Fig.~\ref{fig:budget}) are deployed to 3D cross-domain scenarios, their results are unsatisfactory. Here, we analyze why these methods are not applicable to the cross-domain 3DOD task. We attribute the reason to the following two aspects. \textit{1) 2D Density vs. 3D Sparsity:} Compared with 2D images, 3D point cloud is extremely sparse, which makes Globel Average Pooling (GAP) based feature extractor not suitable for 3D scenes. As a result, directly leveraging CNN on highly sparse feature maps cannot extract informative features. 
\textit{2) 2D Diversity vs. 3D Correlation:} Unlike 2D natural images that have more diverse appearances, the point cloud objects in autonomous driving are closely related, especially between adjacent frames in the same sequence. Thus, simply applying 2D ADA methods to 3DOD will yield similar importance metric scores of candidate data, resulting in labeling redundancy.

\begin{table}[]
\small
\centering
\setlength{\tabcolsep}{1.6mm}{
\begin{tabular}{ccccc}
\hline
\multirow{2}{*}{Source} & \multirow{2}{*}{Target} & \multirow{2}{*}{SN} &  nuScenes$\rightarrow$KITTI & Waymo$\rightarrow$KITTI \\
    & & &  $\text{AP}_{\text{BEV}}$ / $\text{AP}_{\text{3D}}$ & $\text{AP}_{\text{BEV}}$ / $\text{AP}_{\text{3D}}$ \\ 
\hline
- & - & - & 68.15 / 37.17 & 61.18 / 22.01 \\
Ran. & - & \XSolidBrush & 58.02 / 31.09 & 57.49 / 8.78 \\ 
Act. & - & \XSolidBrush & 73.90 / 43.02 & 68.27 / 28.53 \\
Ran. & - & \Checkmark & 70.13 / 58.80 & 71.23 / 56.20 \\ 
Act. & - & \Checkmark & 81.84 / 65.40 & 81.53 / 67.41 \\
Ran. & Ran. & \Checkmark & 84.42 / 75.12 & 85.48 / 75.89  \\
Act. & Ran. & \Checkmark & 85.02 / 75.43 & 85.70 / 76.12 \\
Ran. & Act. & \Checkmark & 86.53 / 76.54 & 86.12 / 76.92 \\
Act. & Act. & \XSolidBrush & 84.91 / 71.56 & 85.13 / 71.36 \\
Act. & Act. & \Checkmark & \textbf{87.00} / \textbf{77.55} & \textbf{87.12} / \textbf{78.03} \\
\hline
\end{tabular}}
\vspace{-0.15cm}
\caption{Component-level ablation studies. Ran. represents random sampling and Act. represents active sampling using our method. The ablation studies are conducted under $1\%$ target annotation budget on PV-RCNN.}
\label{tab:ablation}
\end{table}

\noindent\textbf{Ablation Studies.} 
The effectiveness of two key components, including domainness-aware source sampling strategy and diversity-based target sampling strategy, is verified on nuScenes$\rightarrow$KITTI and Waymo$\rightarrow$KITTI settings. On one hand, Table~\ref{tab:ablation} indicates that the sampling strategy designed for the source domain has a better performance than the random sampling strategy. This is mainly due to that we pick up a portion of frames whose distribution characteristics are similar to the target domain. 
On the one hand, the sampled source data is also beneficial to improve the detector's adaptability, further helping to select more representative samples from the target domain, as verified by comparing Source+Act and Source+Ran in Table~\ref{tab:ablation}. Moreover, Table~\ref{tab:ablation} also shows that the designed diversity-based target domain sampling strategy also can significantly boost the model transferability between domains.

\begin{table}[]
\small
\centering
\resizebox{\linewidth}{!}
{
\begin{tabular}{ccccc}
\hline
\multirow{2}{*}{Score} & \multirow{2}{*}{Entropy} & \multirow{2}{*}{Dom.} & nuScenes$\rightarrow$KITTI & Waymo$\rightarrow$KITTI \\
 &   & &  $\text{AP}_{\text{BEV}}$ / $\text{AP}_{\text{3D}}$ & $\text{AP}_{\text{BEV}}$ / $\text{AP}_{\text{3D}}$ \\ 
\hline
 - & - & S & 79.65 / 58.74 & 77.71 / 55.52 \\
 \Checkmark & - & S & 77.93 / 63.52 & \textbf{82.29} / 54.74 \\ 
 - & \Checkmark & S & 78.90 / 60.02 & 79.79 / 58.43\\
 \Checkmark & \Checkmark & S & \textbf{81.84} / \textbf{65.40} & 81.53 / \textbf{67.41} \\ 
\hline
 - & - & S+T & 86.13 / 76.65 & 85.58 / 76.69 \\
 \Checkmark & - & S+T & 86.09 / 77.21 & 86.71 / 77.91 \\ 
 - & \Checkmark & S+T & 86.39 / 76.82 & 85.91 / 77.96 \\
 \Checkmark & \Checkmark & S+T & \textbf{87.00} / \textbf{77.55} & \textbf{87.12} / \textbf{78.03} \\
 \hline
\end{tabular}
}
\vspace{-0.15cm}
\caption{Ablation study of foreground region-aware discriminator under $1\%$ target annotation budget. Score and Entropy in this table denote that we employ the objectness and entropy evaluation metrics, respectively. S denotes that only the source domain is used for the designed active sampling strategy and S+T represents our Bi3D).}
\label{tab:abl_discriminator}
\vspace{-0.40cm}
\end{table}

As mentioned in Section \ref{subsec:method}, we enhance the scene-level foreground features by combining objectness and entropy scores. In order to verify the necessity of such a design, we conduct experiments by changing the input features of the domain discriminator. It can be observed from Table~\ref{tab:abl_discriminator} that, combining objectness and entropy scores achieves the best target-domain accuracy in both cases of only sampling source data, and sampling source and target data. This shows that the objectness score and entropy score can provide the location information of the foreground and make the model ignore the noisy background. 

\begin{table}[]
\small
\centering
\setlength{\tabcolsep}{1.6mm}{
\begin{tabular}{cccc}
\hline
\multirow{2}{*}{Method} & Source & Target & Avg.\\
 & $\text{AP}_{\text{BEV}}$ / $\text{AP}_{\text{3D}}$ &  $\text{AP}_{\text{BEV}}$ / $\text{AP}_{\text{3D}}$ & $\text{AP}_{\text{BEV}}$ / $\text{AP}_{\text{3D}}$ \\ 
\hline
Source Only & \textbf{47.57} / \textbf{32.43}  & 68.15 / 38.17 & 57.86 / 35.30 \\
ST3D~\cite{yang2021st3d} & 28.57 / 19.88 & 78.36 / 70.85 & 53.46 / 45.36 \\ 
Ours & 40.71 / 22.89 & \textbf{84.91} / \textbf{71.56} & \textbf{62.81} / \textbf{47.22} \\
\hline
\end{tabular}}
\vspace{-0.15cm}
\caption{Generalization ability of Bi3D. We report the results tested on both source and target domains. Avg. denotes the average accuracy across two domains.}
\label{tab:abl_generalization}
\end{table}

\noindent\textbf{Bi3D for Enhancing 3DOD Generalization.} Although domain adaptation tasks including UDA and ADA can achieve higher detection accuracy on a new target domain, their detection accuracy on the original source domain usually degrades after the domain adaptation process is finished. It can be seen from Table~\ref{tab:abl_generalization} that, the source-domain performance achieved by an adapted ST3D will cause a serious performance drop. In contrast, since our Bi3D utilizes a Bi-domain active sampling strategy to pick up both source and target samples, the adapted detector can maintain a certain degree of transferability toward the original source domain. From Table~\ref{tab:abl_generalization}, we can observe that compared with ST3D, our model can obtain a better source domain performance. This shows that our Bi3D can enhance the detector's dataset-level generalization ability for both source and target domains.

\begin{table}[]
\small
\centering
\setlength{\tabcolsep}{1.8mm}{
\begin{tabular}{ccc}
\hline
\multirow{2}{*}{Method} & nuScenes$\rightarrow$KITTI & Waymo$\rightarrow$KITTI \\
 & $\text{AP}_{\text{BEV}}$ / $\text{AP}_{\text{3D}}$ &  $\text{AP}_{\text{BEV}}$ / $\text{AP}_{\text{3D}}$  \\ 
\hline
ST3D~\cite{yang2021st3d} &  84.29 / 72.94 & 86.65 / 76.86 \\
Ours & 87.00 / 77.55  & 87.12 / 78.03 \\ 
Ours+ST3D~\cite{yang2021st3d} &  \textbf{89.28} / \textbf{79.69}  & \textbf{87.83} / \textbf{81.23}\\
Oracle &  88.98 / 82.50   & 88.98 / 82.50 \\
\hline
\end{tabular}}
\vspace{-0.15cm}
\caption{The studies of combining Bi3D with UDA method under $1\%$ target annotation budget on PV-RCNN.}
\label{tab:abl_uda}
\vspace{-0.40cm}
\end{table}

\noindent\textbf{Combining with UDA Method.} Current UDA works~\cite{yang2021st3d} mainly leverage self-training to perform the pseudo-labeling on the unlabeled target domain, which is orthogonal to our Bi3D. Therefore, we conduct the experiments of combining our Bi3D and ST3D~\cite{yang2021st3d}. In particular, we employ Bi3D to actively sample $1\%$ target data to perform the manual annotation, and utilize ST3D to pseudo-label the remaining unlabeled target data. Then, we fine-tune the detector using both annotated data and pseudo-labeled data. It can be seen from Table \ref{tab:abl_uda} that, our Bi3D can be flexibly combined with ST3D, significantly outperforming both the Bi3D and ST3D methods.


\section{Conclusion}
In this work, for the first time, we presented a Bi3D framework, which develops a Bi-domain active sampling approach to dynamically select important frames from both source and target domains, achieving domain transfer at a low data cost. Experimentally, Bi3D achieves consistent accuracy gains on many cross-domain settings, \eg, for Waymo-to-KITTI setting, Bi3D re-trained on only $5\%$ target domain data (KITTI) outperforms the corresponding baseline model trained using $100\%$ labeled KITTI data. 

\section*{Acknowledgement}
This work is supported by National Natural Science Foundation of China (No. U1909207 and 62071127),  Zhejiang Lab Project (No. 2021KH0AB05) and Science and Technology Commission of Shanghai Municipality (grant No. 22DZ1100102).


{\small
\bibliographystyle{ieee_fullname}
\bibliography{egbib}
}

\clearpage
\appendix

\crefname{section}{Sec.}{Secs.}
\Crefname{section}{Section}{Sections}
\Crefname{table}{Table}{Tables}
\crefname{table}{Tab.}{Tabs.}

\def\cvprPaperID{7154} 
\def\confName{CVPR}
\def\confYear{2023}


\centerline{\large{\textbf{Outline}}}
Due to the eight-page limitation of the submission paper, we provide more details and visualizations from the following aspects:
\begin{compactitem}
    \setlength{\itemsep}{1.0pt}
    \setlength{\parsep}{1.0pt}
    \setlength{\parskip}{1.0pt}
    \item Sec.~\ref{sec:Methods}: More details of the proposed Bi3D.
    \item Sec.~\ref{sec:Datasets}: Details of datasets.
    \item Sec.~\ref{sec:implementation_detail}: Implementation details for Bi3D and 2D image-related ADA works.
    \item Sec.~\ref{sec:experiments}: More experimental results.
    \item Sec.~\ref{sec:qualitative_results}: Qualitative results. 
\end{compactitem}

\section{More details of Bi3D}
\label{sec:Methods}

\subsection{Detailed Structure of Domain Discriminator}
As mentioned in the Method Section in our main text, we use a conventional convolution-based discriminator to measure the domainness score of samples from source and target domains. In particular, given a BEV feature map {\small $\hat{f}_{bev}^d \in \mathbb{R}^\mathrm{C\times H\times W}$} enhanced by objectness and entropy scores from source or target domain, where {\small $d\in [s,t]$} denotes source or target domain and {\small $H$}, {\small $W$}, {\small $C$} represent the channel, height and width of the feature map, respectively. As shown in Fig. \ref{fig:discriminator}, we first use 5 layers convolution followed by LeakyReLU to extract scene-level features and the features are down-sampled by 32 times in this process. Then, the Global Average Pooling (GAP) is utilized to extract the scene-level vectors. Further, we use a single layer Multi-Layer Perception (MLP) to obtain the domainness score of each candidate frame. 

\setcounter{figure}{3}
\begin{figure}[h]
\centering
\includegraphics[width=4.0cm]{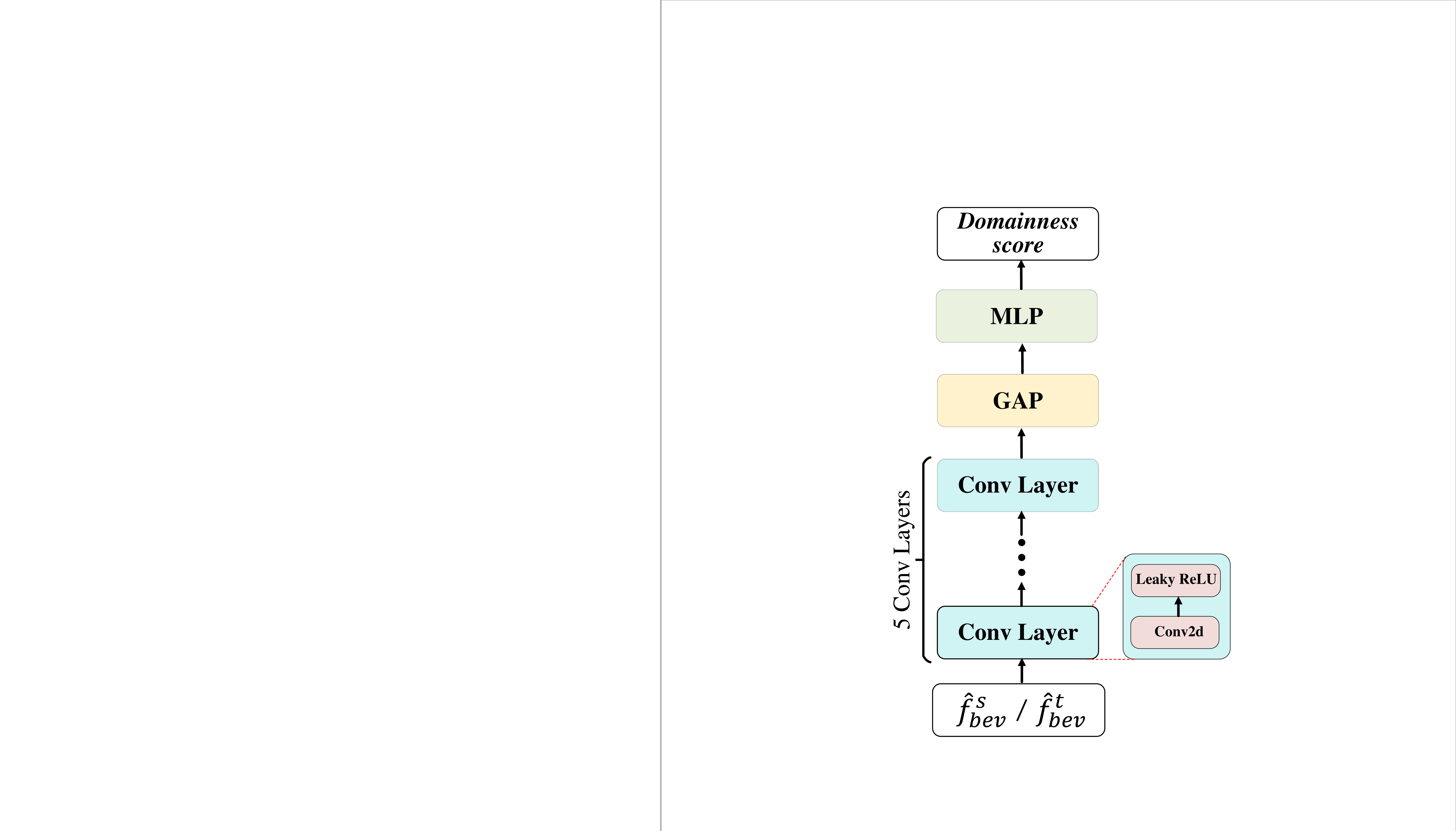}
\caption{Detailed structure of domain discriminator. $\hat{f}_{bev}^s$ and $\hat{f}_{bev}^t$ denote enhanced foreground features from source domain and target domain.}
\label{fig:discriminator}
\end{figure}

\subsection{Algorithm of the Proposed Bi3D}
In this Section, we show the detail training procedure of the Bi3D in Algorithm \ref{alg:bi3d}. 

\begin{algorithm}[h]
	\caption{Bi3D}
	\label{alg:bi3d}
	\begin{algorithmic}[1]
		\State Train the detector on train dataset $D_s$.
		\State Train the domain discriminator on $D_s$ and $D_t$.
		\State Select target-domain-like source data $\tilde{D}_s$ by domainness-aware source sampling strategy.
		\State Fine-tune the detector on $\tilde{D}_s$.
		\For{current epoch $<$ max epochs}
			\If{current epoch in sample epochs}
				\State Select diverse and representative target data \State  $\Delta D_t$ using diversity-based target sampling \State  strategy and update $\tilde{D}_t$.
			\EndIf
			\State Train detector on $\tilde{D}_t$ and $\tilde{D}_s$.
		\EndFor
	\end{algorithmic}
\end{algorithm}

\section{Datasets}
\label{sec:Datasets}

\noindent\textbf{KITTI.} KITTI dataset~\cite{geiger2012we} is one of the most popular datasets for autonomous driving, which contains 7481 training samples and is divided into a train set with 3712 samples and a validation set with 3769 samples. The point cloud data is collected by a 64-beam LiDAR in Germany.  Due to only annotations of the field of view of the front (FOV) camera is provided, we remove the points outside of the range in the test phase.  

\noindent\textbf{Waymo.} Waymo Open Dataset~\cite{sun2020scalability} is a large-scale autonomous driving dataset which is composed of 1000 sequences and divided into a train set with 798 sequences ({\small $\sim$}1.5 million samples) and a validation set with 202 sequences ({\small $\sim$}4 million samples). The Waymo dataset is gathered in the USA by a 64-beam LiDAR with annotations in full $360^\circ$. We use one fifth data of Waymo train set when Waymo is regarded to the source domain.

\noindent\textbf{nuScenes.} nuScnenes dataset~\cite{caesar2020nuscenes} provides point cloud data from 32-beam LiDAR collected from Singapore and Boston, USA. It consists of 28130 training samples and 6019 validation samples. The data is obtained during different time in a day and different weathers.

\noindent\textbf{Lyft.} Lyft level 5 dataset~\cite{lyft2019} is composed of 18900 training samples and 3780 validation samples, which are collected in the USA. As mentioned in ST3D~\cite{yang2021st3d}, the Lyft dataset does not annotate objects on both sides of the road which is different from other datasets and will cause the detection accuracy degradation.

\section{Implementation Details}
\label{sec:implementation_detail}

\setcounter{table}{5}
\subsection{More Bi3D Implementation Details}
\label{Bi3D_implementation}
\begin{table}[h]
	\centering
	\setlength{\tabcolsep}{2.5mm}{
		\begin{tabular}{c|c}
			\hline
			Dataset & Point Could Range \\
			\hline
			KITTI~\cite{geiger2012we} & $[0, -40, -3, 70.4, 40, 1]$ \\
			Waymo~\cite{sun2020scalability} &  $ [-75.2, -75.2, -2, 75.2, 75.2, 4]$ \\
			nuScenes~\cite{caesar2020nuscenes} & $[-51.2, -51.2, -5.0, 51.2, 51.2, 3.0]$ \\
			Lyft~\cite{lyft2019} & $[-80.0, -80.0, -5.0, 80.0, 80.0, 3.0]$\\
			\hline
	\end{tabular}}
	\caption{Point cloud range of different datasets.}
	\label{tab:range}
\end{table}

\begin{table}[h]
	\centering
	\setlength{\tabcolsep}{1.5mm}{
		\begin{tabular}{ccccc}
			\hline
			& Budget & KITTI & nuScenes & Lyft \\
			\hline
			number & \multirow{2}{*}{$1\%$}  & $18$ & $140$ & $94$  \\
			epochs & & $[0,5]$ & $[0,5]$ & $[0,5]$ \\
			\hline
			number & \multirow{2}{*}{$5\%$} & $37$ & $280$ & $188$ \\ 
			epochs &  & $[0,2,4,6,8]$ & $[0,2,4,6,8]$ & $[0,2,4,6,8]$ \\
			\hline
	\end{tabular}}
	\caption{Details of target sample budget and sample epochs. Number denotes the sampling number per epoch.}
	\label{tab:abl_imp}
\end{table}

As shown in Table \ref{tab:range}, the point cloud range varies in different datasets, followed by ST3D~\cite{yang2021st3d}, we align the point cloud range with the Waymo dataset (\textit{i.e.,} {\small $ [-75.2, -75.2, -2, 75.2, 75.2, 4]$}). In active sampling target domain stage, when annotation budget {\small $B$} of the target domain is equal to $1\%$, we perform a two-round sampling process. Besides, the sampling process will be performed $5$ rounds, when annotation budget $B$ is set to $5\%$. The sample number and sample epochs of the active sampling target data is shown in Table \ref{tab:abl_imp}. The widely-used data augmentation methods (\eg, random world flip, random world rotation, random world scaling) are also used in our experiments. 

\subsection{Details about Reproduced 2D ADA Methods}

\noindent\textbf{TQS}~\cite{fu2021transferable}\textbf{.} TQS consists of transferable committee, transferable uncertainty and transferable domainness which is based on image-level features. In our reproduction, we extract the scene-level feature from BEV features using CNN and use three classifier heads to construct the committee. Following TQS~\cite{fu2021transferable}, different from previous works using entropy to evaluate the uncertainty, we calculate the the margin of objectness score and $0.5$ which is also different from the original TQS. This is because we only focus on a single category and cannot use margin of the highest score and the second-highest score. In addition, we calculate domainness score by a domain discriminator and set $\mu$ as 0.75 and $\sigma$ as 0.4, which is the same as TQS. In order to keep consistency with TQS, we use source data and selected target data to fine-tune the detector trained on the source domain.

\noindent\textbf{CLUE}~\cite{prabhu2021active}\textbf{.} CLUE uses uncertainty-weighted cluster to select target data. Following CLUE, we first obtain the uncertainty of each frame by calculating the predictive entropy of the proposals after NMS and then use the mean value of the entropy to represent the frame-level uncertainty. Further, weighted K-Means which is proposed by CLUE with $b_k$ centroids is utilized to cluster, where $b_k$ denotes the annotation budget of the current sampling epoch.

The sampling number and sampling epochs are consistent with our Bi3D, as shown in Table \ref{tab:abl_imp}.

\subsection{Details about reproduced 2D AL Methods}

\noindent\textbf{Query by Committee}\cite{Schohn2000LessIM}\textbf{.} Committee-based methods often use multiple classifier heads with different initialization to keep diversity from each other. We use two classifiers with different initialization and calculate the distance of output logits of two classifier heads and select the target data which the two classifier heads most disagree.  

\noindent\textbf{Query by Uncertainty}\cite{wang2014new}\textbf{.} The most common method to measure the uncertainty is to calculate the entropy. Here, we calculate the entropy of the instance-level classifier score of the proposals after NMS and use the mean value of the entropy to represent the frame-level uncertainty.

The sampling number and sampling epochs are consistent with our Bi3D, as shown in Table \ref{tab:abl_imp}.

\section{Experimental Results}
\label{sec:experiments}

\subsection{Second-IOU Results}
SECOND~\cite{yan2018second} is a widely used detector that greatly improves the efficiency of the model by using sparse convolution. Followed by~\cite{yang2021st3d,wei2022lidar}, we also conduct the experiment on SECOND-IoU, where an extra IoU head is added to SECOND. As shown in Table \ref{tab:second}, the proposed Bi3D can also achieve the best result compared to UDA methods by only sampling $1\%$ data from nuScenes. This shows our approach is applicable to multiple detectors.

\begin{table}[h]
	\centering
	\setlength{\tabcolsep}{1.5mm}{
		\begin{tabular}{c|c|c}
			\hline
			Task & Method &  $\text{AP}_\text{BEV}$ / $\text{AP}_\text{3D}$  \\
			\hline
			
			\multirow{5}{*}{Waymo$\rightarrow$nuScenes} &  Source Only  & 32.91 / 17.24 \\
			& SN & 33.23 / 18.57 \\
			& ST3D & 35.92 / 20.19 \\ 
			& Lidar Distill & 40.66 / 22.86 \\
			& Ours & \textbf{42.15} / \textbf{26.24}\\
			\hline
	\end{tabular}}
	\caption{Results of Waymo$\rightarrow$nuScenes adaptation task using SECOND-IoU. Source Only means that the model trained on the Waymo dataset is directly tested on nuScenes. We report $\text{AP}_\text{BEV}$ and $\text{AP}_\text{3D}$ over 40 recall positions of the car category at IoU = 0.7.}
	\label{tab:second}
\end{table}

\subsection{More Ablation Studies}

\begin{table}[h]
	\centering
	\setlength{\tabcolsep}{4mm}{
		\begin{tabular}{c|c|c}
			\hline
			\multicolumn{2}{c}{Number of Source} &  $\text{AP}_\text{BEV}$ / $\text{AP}_\text{3D}$  \\
			\hline
			
			\multirow{5}{*}{Proportion}& $1\%$  & 86.95 / \textbf{77.86} \\
			& $2\%$ & 86.55 / 77.55 \\
			& $5\%$ & 86.66 / 77.59 \\ 
			& $10\%$ & 85.89 / 76.78 \\
			& $25\%$ & 85.24 / 75.68 \\
			\hline
			Threshold & $>0$ & \textbf{87.00} / 77.58\\
			\hline
	\end{tabular}}
	\caption{Experiments on number of selected source data. We conduct the experiments on Waymo$\rightarrow$nuScenes adaptation task using PV-RCNN. Number of Source indicates the number of selected source data using domainness-aware source sample strategy. We report $\text{AP}_\text{BEV}$ and $\text{AP}_\text{3D}$ over 40 recall positions of the car category at IoU = 0.7.}
	\label{tab:source}
\end{table}

\noindent\textbf{Experiments on number of selected source data.} In our main text, the experimental results have shown the effectiveness of our proposed domainness-aware source sampling strategy. To fully explore the influence of the number of selected source data, we conduct further experiments. As shown in Table \ref{tab:source}, selecting more source domain data degrades performance (\eg, select $25\%$ source data) as the domain gap between {\small $\tilde{D}_s$} and {\small $D_t$} becomes larger, where {\small $\tilde{D}_s$} denotes the set of selected target-domain-like source data and {\small $D_t$} represents target dataset.

\begin{figure}
	
	\begin{subfigure}{0.49\linewidth}
		\includegraphics[width=4.1cm]{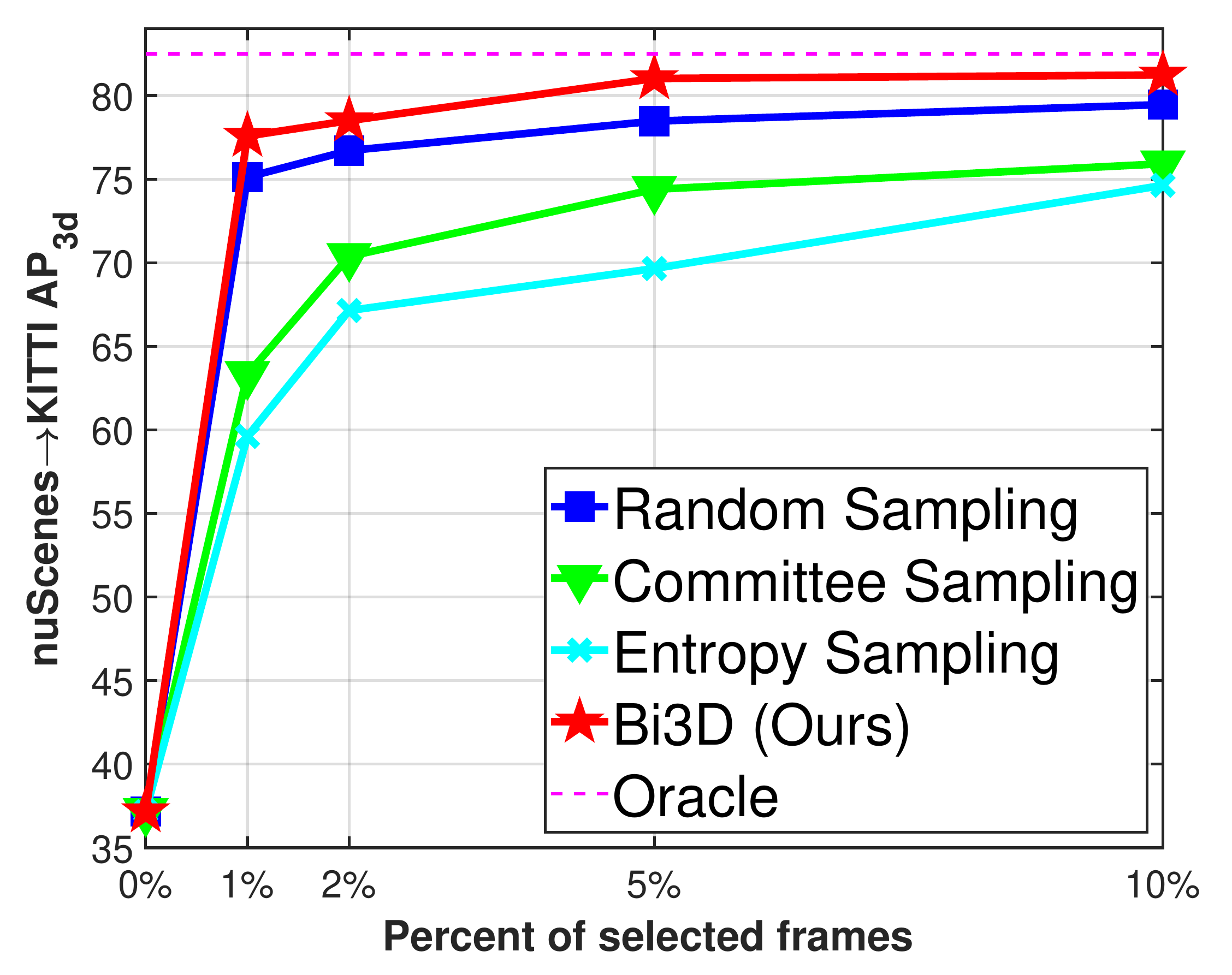}
		\caption{nuScenes$\rightarrow$KITTI}
		\label{fig:subfig_1}
	\end{subfigure}
	\begin{subfigure}{0.49\linewidth}
		\includegraphics[width=4.1cm]{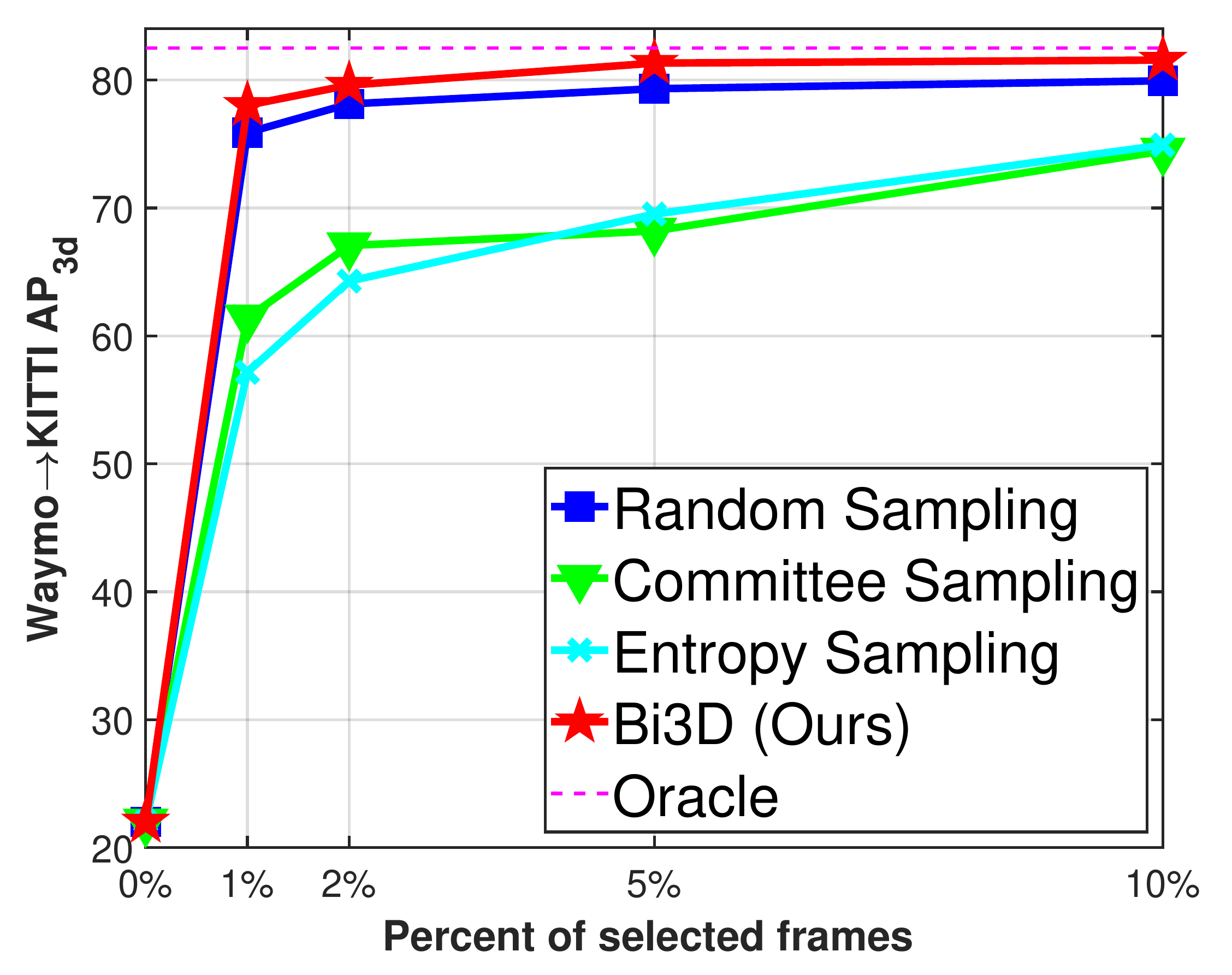}
		\caption{Waymo$\rightarrow$KITTI}
		\label{fig:subfig_2}
	\end{subfigure}
	\caption{Results of various target annotation budgets.}
	\label{fig:budget_3d}
\end{figure}

\noindent\textbf{Varying Annotation Budget.} We show the line chart of $\text{AP}_\text{BEV}$ at different target annotation budgets in our main text. Here we show the trend of $\text{AP}_\text{3D}$. As shown in Fig. \ref{fig:budget_3d}, similar to the results of $\text{AP}_\text{BEV}$, the proposed Bi3D achieves a promising detection accuracy gain, even outperforming many active learning methods. In addition, with the increase of the number of manual annotated target frames, the model detection accuracy is constantly improved.

\subsection{Results with IoU=0.5}
In this section, we show $\text{AP}_\text{BEV}$ and $\text{AP}_\text{3D}$ results with the IoU threshold 0.5. The results are shown in Table \ref{tab:iou5}, we can observe that our Bi3D can also achieve the best performance on various cross-domain 3DOD tasks.
\begin{table*}[htbp]
	\centering
	\begin{small}
		\begin{tabular}{c|c|c|c}
			\bottomrule[1pt]
			\multirow{2}{*}{Task} & \multirow{2}{*}{Method}  &PV-RCNN & Voxel R-CNN \\
			\hhline{~~|-|-}
			&  & $\text{AP}_{\text{BEV}}$ / $\text{AP}_{\text{3D}}$ &  $\text{AP}_{\text{BEV}}$ / $\text{AP}_{\text{3D}}$ \\
			\hline
			\multirow{10}{*}{Waymo$\rightarrow$KITTI} & Source Only  & 88.33 / 87.17 & 89.49 / 87.41 \\
			& ST3D~\cite{yang2021st3d} & 92.40 / \textbf{92.18} &  92.34 / 84.92\\
			& Ours & \textbf{93.88} / \textbf{92.18} & \textbf{92.76} / \textbf{92.08} \\
			\hhline{~|-|-|-}
			& SN \cite{wang2020train} & 86.32 /85.72 & 82.00 / 81.57 \\
			& ST3D (w/ SN)~\cite{yang2021st3d} & 91.49 / 90.77 & 86.76 / 86.40  \\
			& CLUE (w/ SN)~\cite{prabhu2021active} & 90.42 / 88.87 & 88.13 / 88.00 \\
			& TQS (w/ SN)~\cite{fu2021transferable} & 88.17 /89.87 & 84.77 / 83.00 \\
			& Ours (w/ SN) & \textbf{92.48} / \textbf{92.31} & \textbf{94.31} / \textbf{94.16} \\
			& Ours (w/ SN, $5\%$) & \textbf{93.31} / \textbf{93.26} & \textbf{94.27} / \textbf{94.17} \\
			
			\hhline{~|-|-|-}
			& Oracle & 94.08 / 92.28 & 95.54 / 95.50 \\
			\toprule[1pt]
			\bottomrule[1pt]
			\multirow{10}{*}{Waymo$\rightarrow$Lyft} & Source Only  & 82.38 / 80.45 & 77.09 / 75.16 \\
			& ST3D~\cite{yang2021st3d}  & 84.52 / 82.61 & 78.43 / 76.68 \\
			& Ours & \textbf{86.03} / \textbf{83.90} & \textbf{85.04} / \textbf{83.04} \\
			\hhline{~|-|-|-}
			& SN \cite{wang2020train}  & 80.12 / 78.09 & 76.26 / 73.54\\
			& ST3D (w/ SN)~\cite{yang2021st3d} & 82.21 / 81.70 & 77.46 / 75.08\\
			& CLUE (w/ SN)~\cite{prabhu2021active} & 84.00 / 81.96 & 83.74 / 81.77 \\
			& TQS (w/ SN)~\cite{fu2021transferable} & 75.60 / 73.45 & 77.64 / 75.67 \\
			& Ours (w/ SN) & \textbf{86.03} / \textbf{83.86} & \textbf{86.19} / \textbf{83.70}\\
			& Ours (w/ SN, $5\%$) & \textbf{86.70} / \textbf{84.26} & \textbf{86.84} / \textbf{83.58} \\
			\hhline{~|-|-|-}
			& Oracle & 92.38 / 91.87 & 90.19 / 87.18\\
			\toprule[1pt]
			\bottomrule[1pt]
			\multirow{10}{*}{Waymo$\rightarrow$nuScenes} & Source Only  & 40.48 / 36.95 & 31.62 / 28.25 \\
			& ST3D~\cite{yang2021st3d} & 40.90 / 38.67 & 44.06 / 34.62 \\
			& Ours & \textbf{52.99} / \textbf{49.17} & \textbf{52.39} / \textbf{48.83} \\
			\hhline{~|-|-|-|}
			& SN \cite{wang2020train} & 40.27 / 36.59 & 33.99 / 31.23  \\
			& ST3D (w/ SN)  \cite{yang2021st3d} & 41.42 / 38.99 & 38.27 / 34.31 \\
			& CLUE (w/ SN) \cite{prabhu2021active}  & 43.79 / 40.80 & 42.67 / 39.87\\
			& TQS (w/ SN)~\cite{fu2021transferable} & 41.10 / 38.01 & 38.87 / 36.38\\
			& Ours (w/ SN) & \textbf{52.65} / \textbf{48.01} & \textbf{52.73} / \textbf{49.04} \\
			& Ours (w/ SN, $5\%$) & \textbf{55.63} / \textbf{51.78} & \textbf{53.01} / \textbf{49.63} \\
			\hhline{~|-|-|-|}
			& Oracle & 61.52 / 58.04 & 58.33 / 54.61 \\
			\toprule[1pt]
			\bottomrule[1pt]
			\multirow{10}{*}{nuScenes$\rightarrow$KITTI} & Source Only & 80.88 / 78.47 & 85.81 / 81.76 \\
			& ST3D \cite{yang2021st3d} & 83.75 / 83.64 & 92.33 / 82.93 \\
			& Ours & \textbf{92.54} / \textbf{92.36} & \textbf{94.86} / \textbf{93.28}\\
			\hhline{~|-|-|-}
			& SN \cite{wang2020train} & 66.22 / 65.82 & 48.59 / 47.49 \\
			& ST3D (w/ SN) \cite{yang2021st3d} & 90.47 / 90.25 & 80.08 / 78.51 \\
			& CLUE (w/ SN) \cite{prabhu2021active}  & 82.04 / 80.59 &  85.97 / 82.86 \\
			& TQS (w/ SN)~\cite{fu2021transferable} & 91.90 / 90.37 & 85.88 / 84.39 \\
			& Ours (w/ SN) & \textbf{92.93} / \textbf{92.74} & \textbf{93.47} / \textbf{93.32} \\
			& Ours (w/ SN, $5\%$) & \textbf{94.70} / \textbf{93.44} & \textbf{93.65} / \textbf{92.46} \\
			
			\hhline{~|-|-|-}
			& Oracle & 94.97 / 94.85 & 95.54 / 95.50 \\
			\toprule[0.8pt]
		\end{tabular}
	\end{small}
	\caption{Results of different adaptation scenarios under $1\%$ annotation budget. Following ~\cite{yang2021st3d, wei2022lidar}, we report $\text{AP}_{\text{BEV}}$ and $\text{AP}_{\text{3D}}$ over recall 40 positions of the car category at IoU = 0.5. Source Only denotes that the pre-trained detector is directly evaluated on the target domain, and Oracle represents that the detection results using the fully-annotated target domain. Closed Gap denotes the performance gap closed by various approaches along Source Only and Oracle results. The best adaptation results are marked in \textbf{bold}.}
	\vspace{-0.3cm}
	\label{tab:iou5}
\end{table*}

\section{Qualitative Results}
\label{sec:qualitative_results}

\subsection{t-SNE results}

\begin{figure}[t]
\centering
\includegraphics[width=6.0cm]{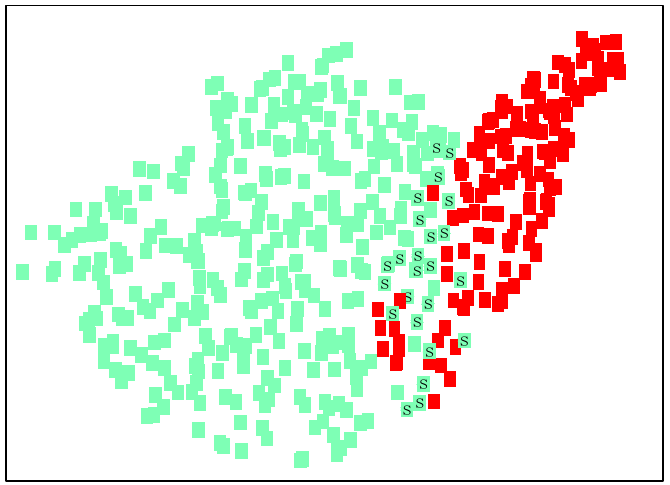}
\caption{t-SNE result of domainness-aware source sampling strategy. The green squares represent data from source domain and red squares represent target domain. We mark the selected source data with 's'.}
\label{fig:source_tsne}
\end{figure}

\begin{figure}
	\begin{subfigure}{0.49\linewidth}
		\includegraphics[width=4.1cm]{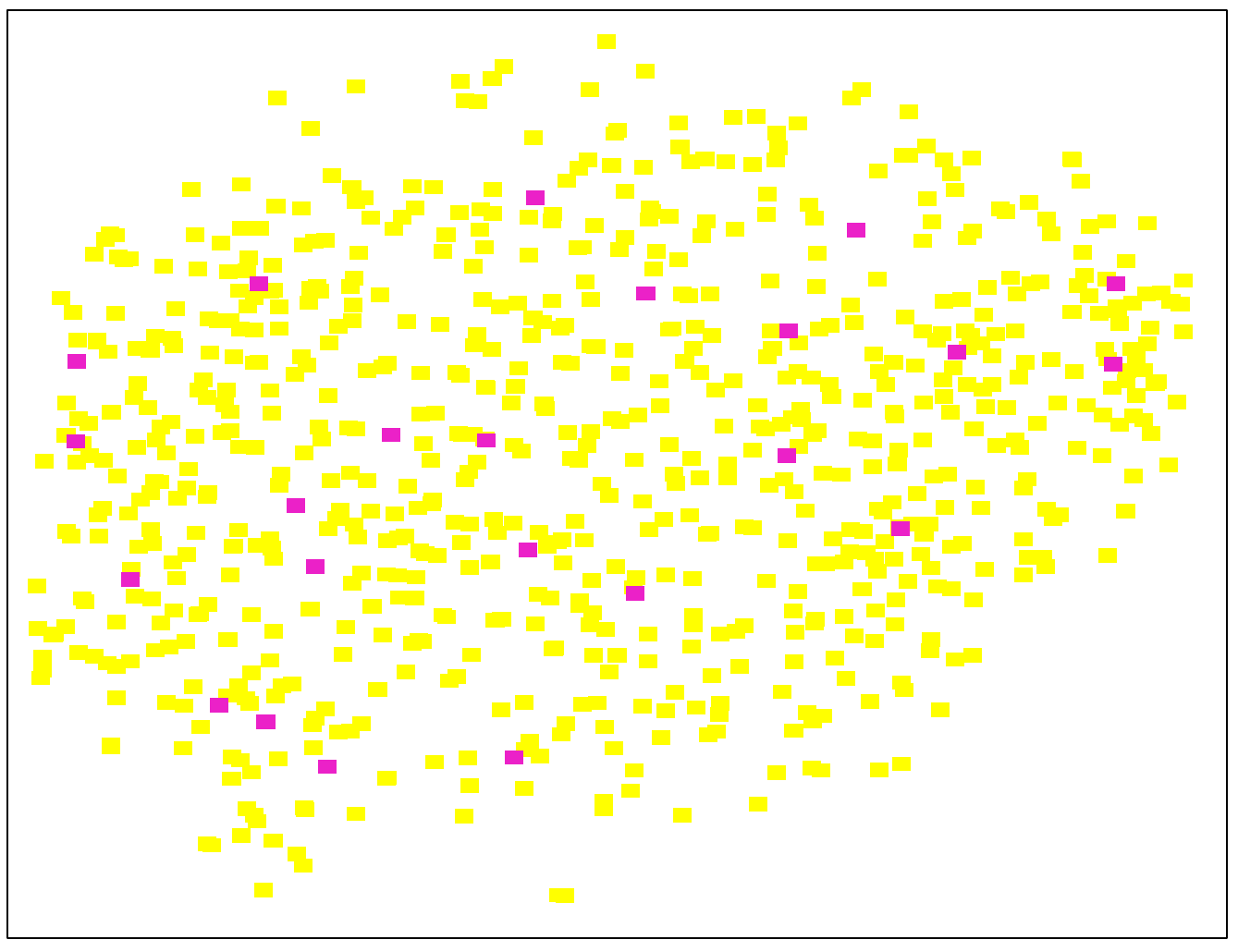}
		\caption{Round 1}
		\label{fig:subfig_a1}
	\end{subfigure}
	\begin{subfigure}{0.49\linewidth}
		\includegraphics[width=4.1cm]{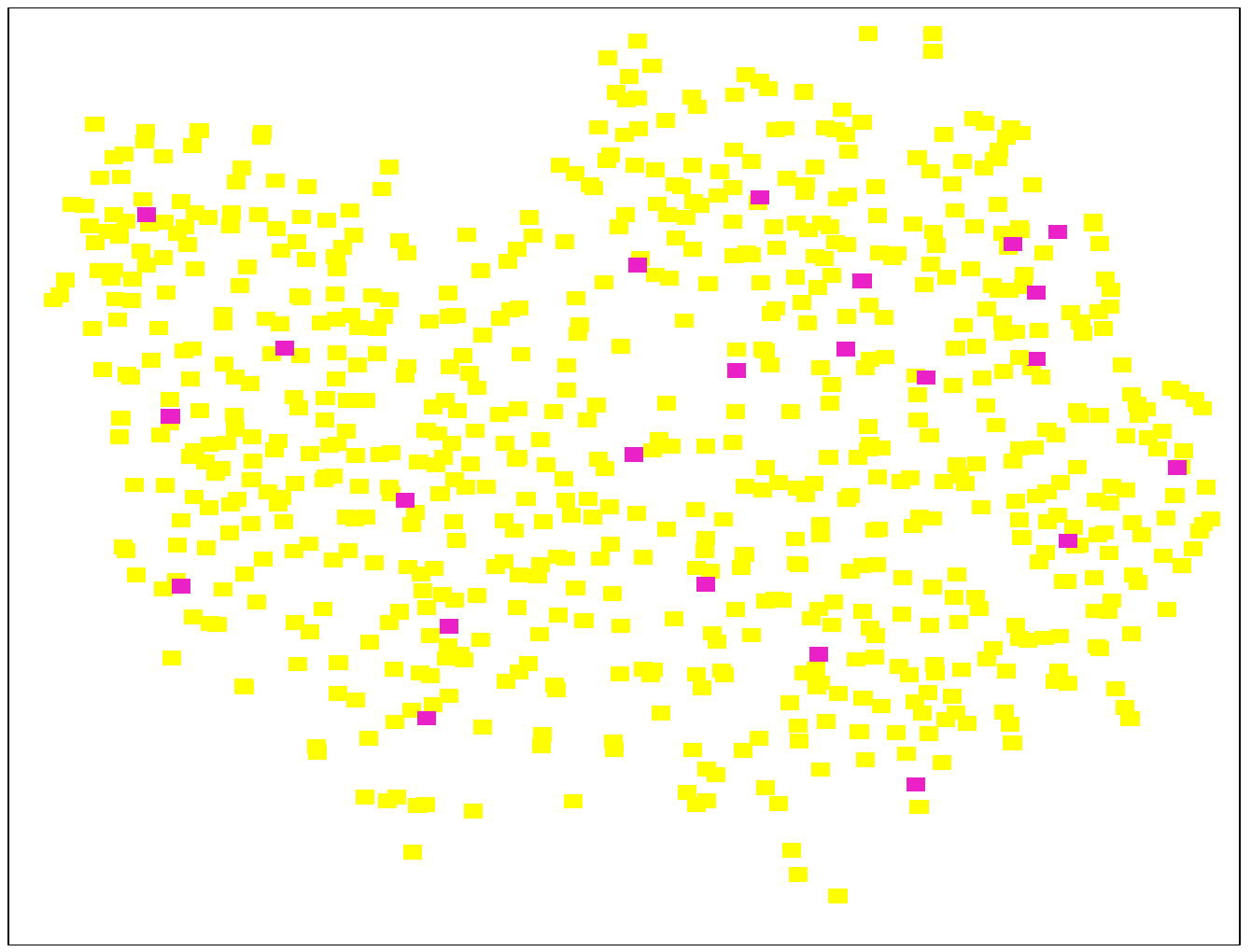}
		\caption{Round 2}
		\label{fig:subfig_b1}
	\end{subfigure}
	\caption{t-SNE result of diversity-based target sampling strategy. The purple squares represent selected target data. As mentioned in \ref{Bi3D_implementation}, we perform a two round sampling strategy. (a) and (b) are the t-SNE results of each sampling round.}
	\label{fig:target_tsne}
\end{figure}

\begin{figure*}[t]
\vspace{-0.40cm}
    \begin{minipage}{0.33\linewidth}
		\centering
		\includegraphics[width=0.9\linewidth]{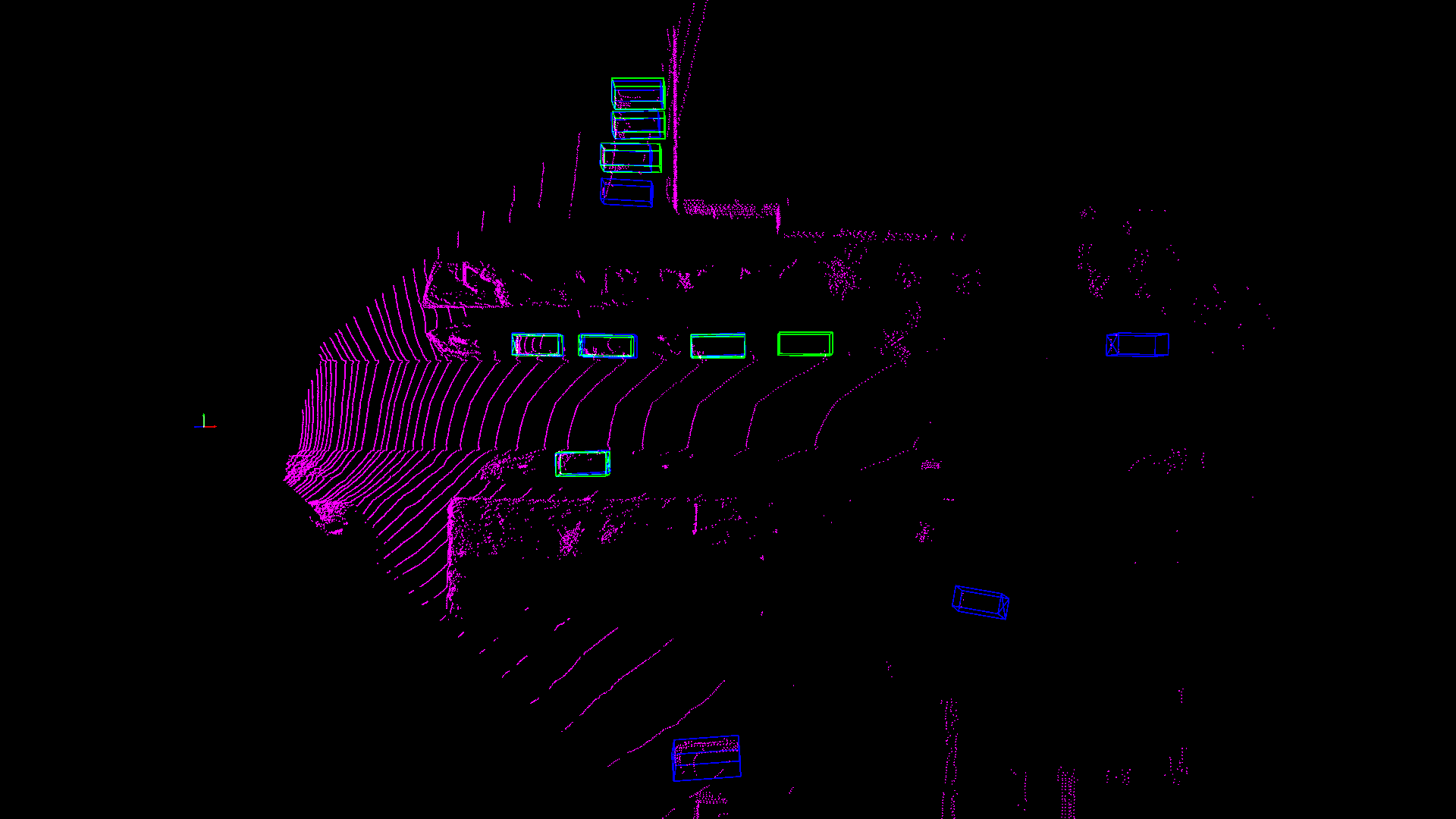}
	\end{minipage}
	\begin{minipage}{0.33\linewidth}
		\centering
		\includegraphics[width=0.9\linewidth]{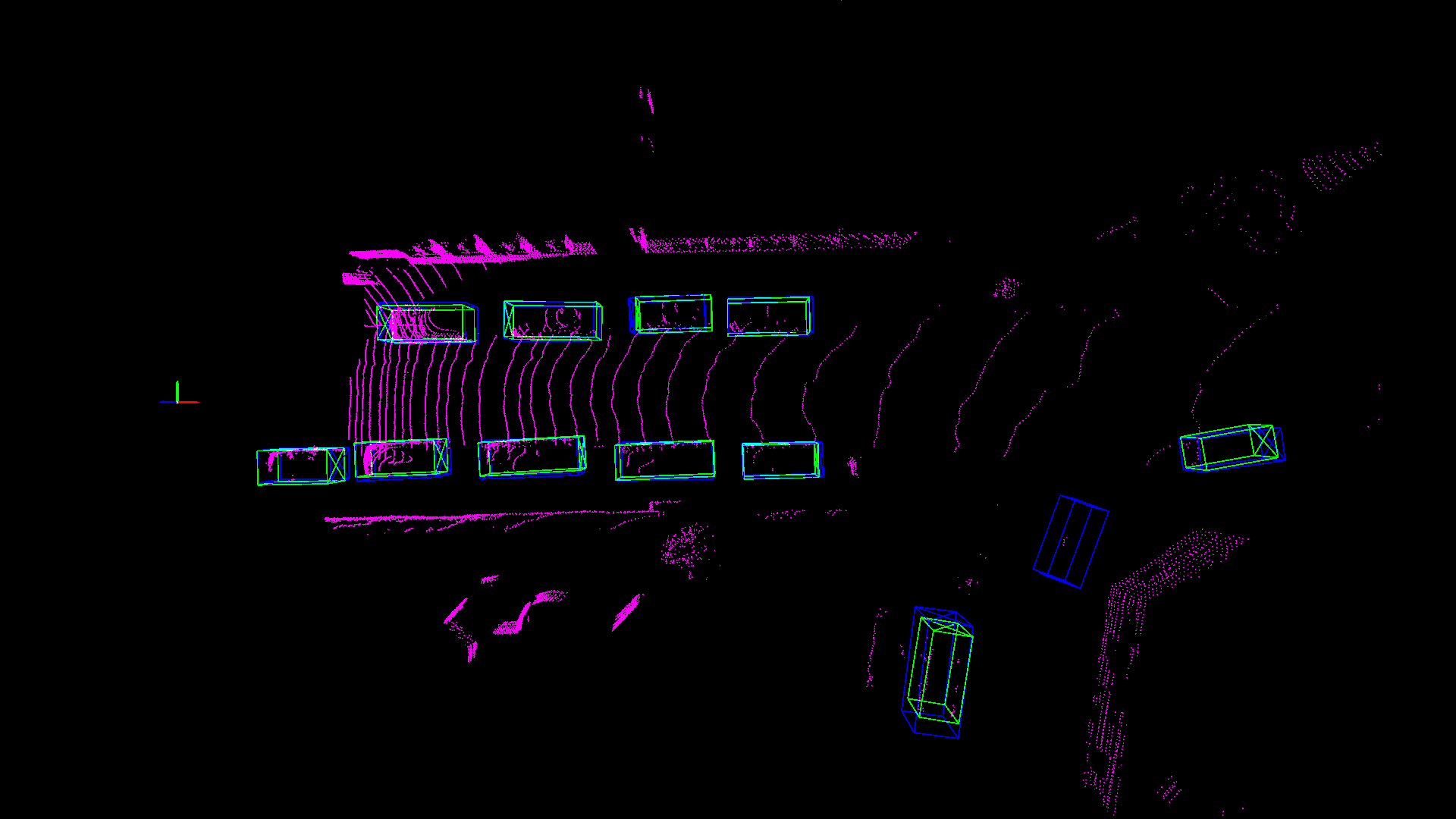}
	\end{minipage}
	\begin{minipage}{0.33\linewidth}
		\centering
		\includegraphics[width=0.9\linewidth]{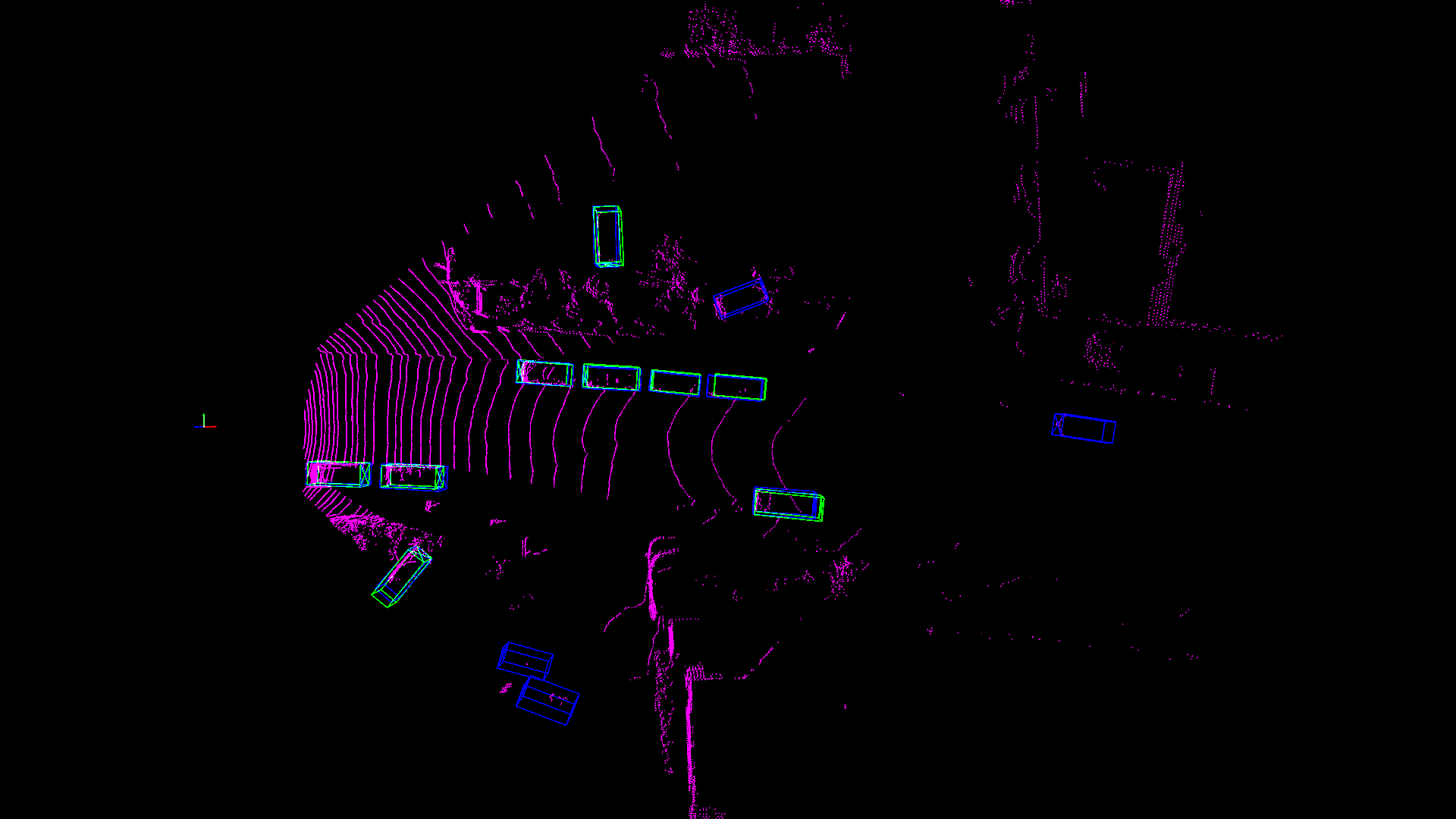}
	\end{minipage}
	\begin{minipage}{0.33\linewidth}
		\centering
		\includegraphics[width=0.9\linewidth]{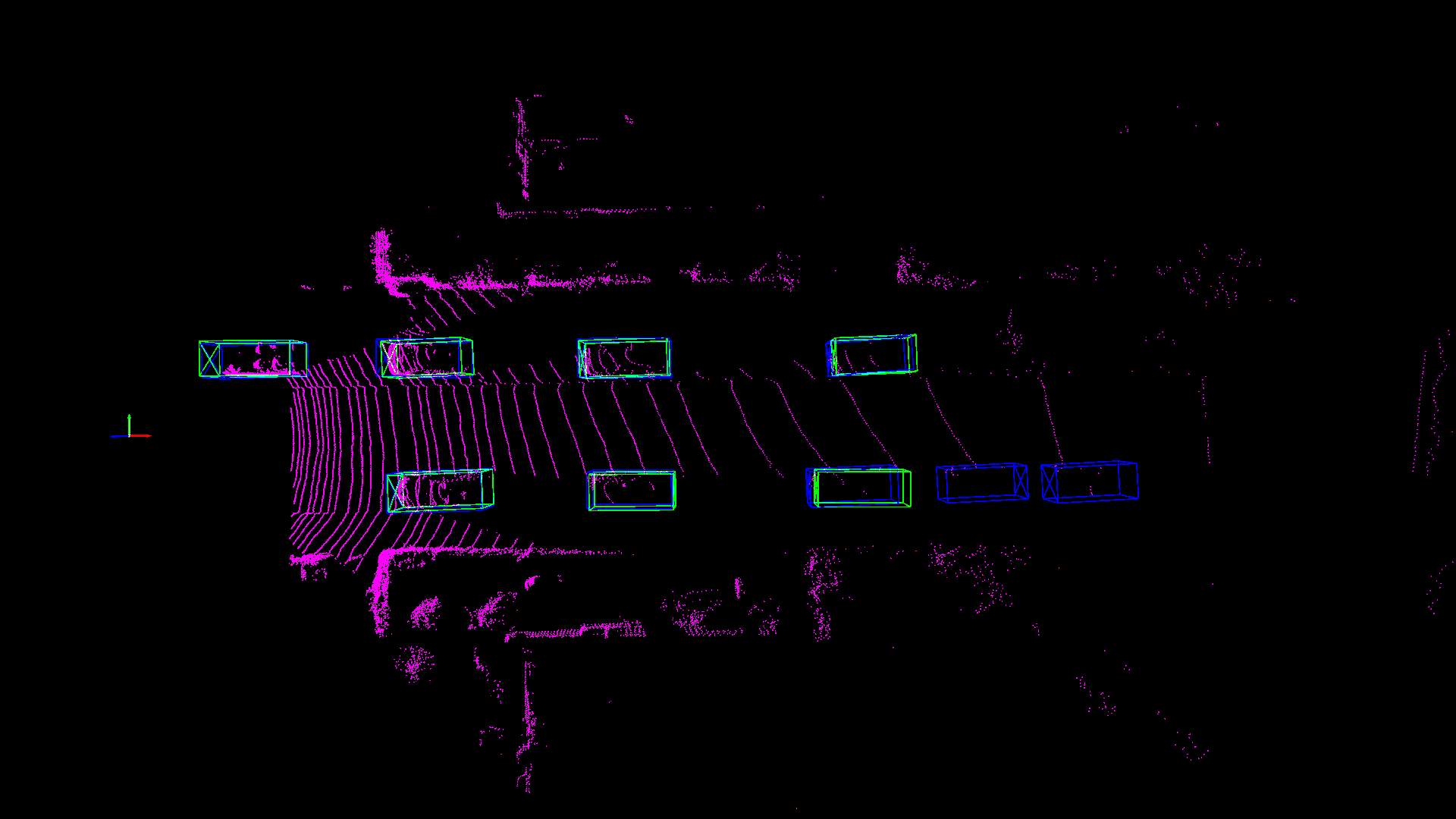}
	\end{minipage}
	\begin{minipage}{0.33\linewidth}
		\centering
		\includegraphics[width=0.9\linewidth]{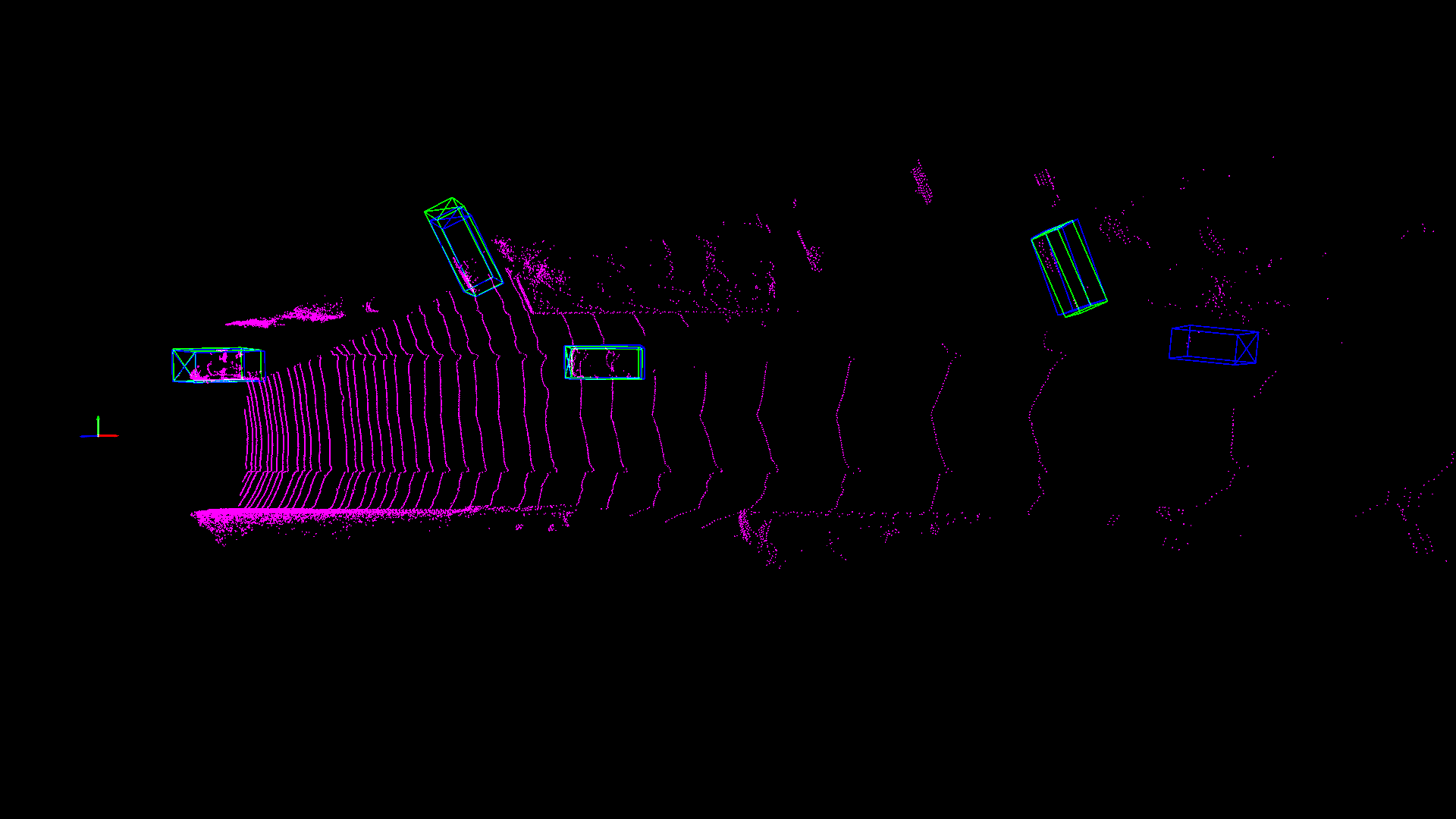}
	\end{minipage}
    \begin{minipage}{0.33\linewidth}
		\centering
		\includegraphics[width=0.9\linewidth]{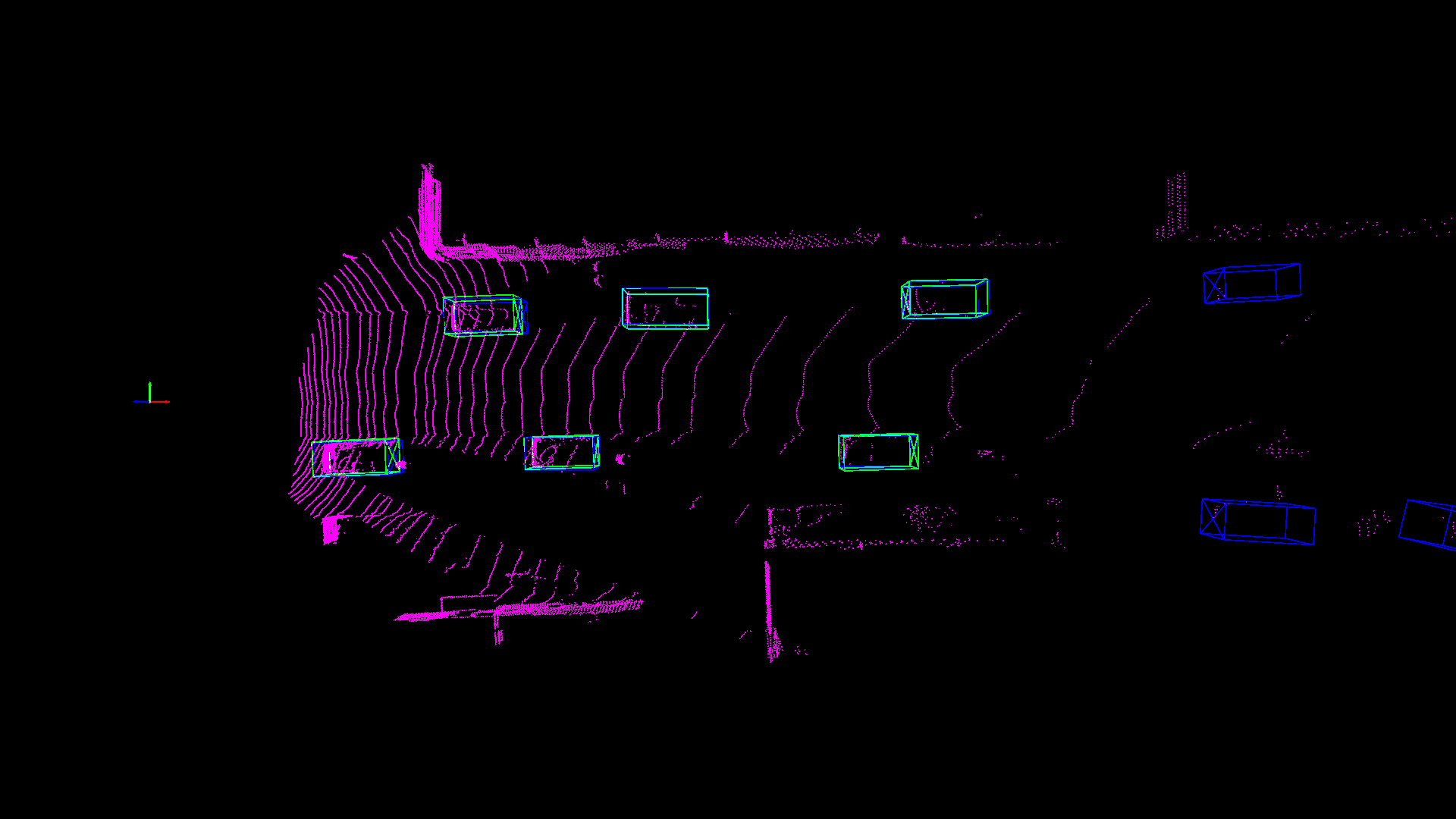}
	\end{minipage}
	\caption{Qualitative results of Waymo-to-KITTI cross-domain scenario. The green and blue bounding boxes represent groundtruths and detector predictions respectively.}
	\label{fig:kitti_with_gt}
\end{figure*}

\begin{figure*}[h]
    \begin{minipage}{0.33\linewidth}
		\centering
		\includegraphics[width=0.9\linewidth]{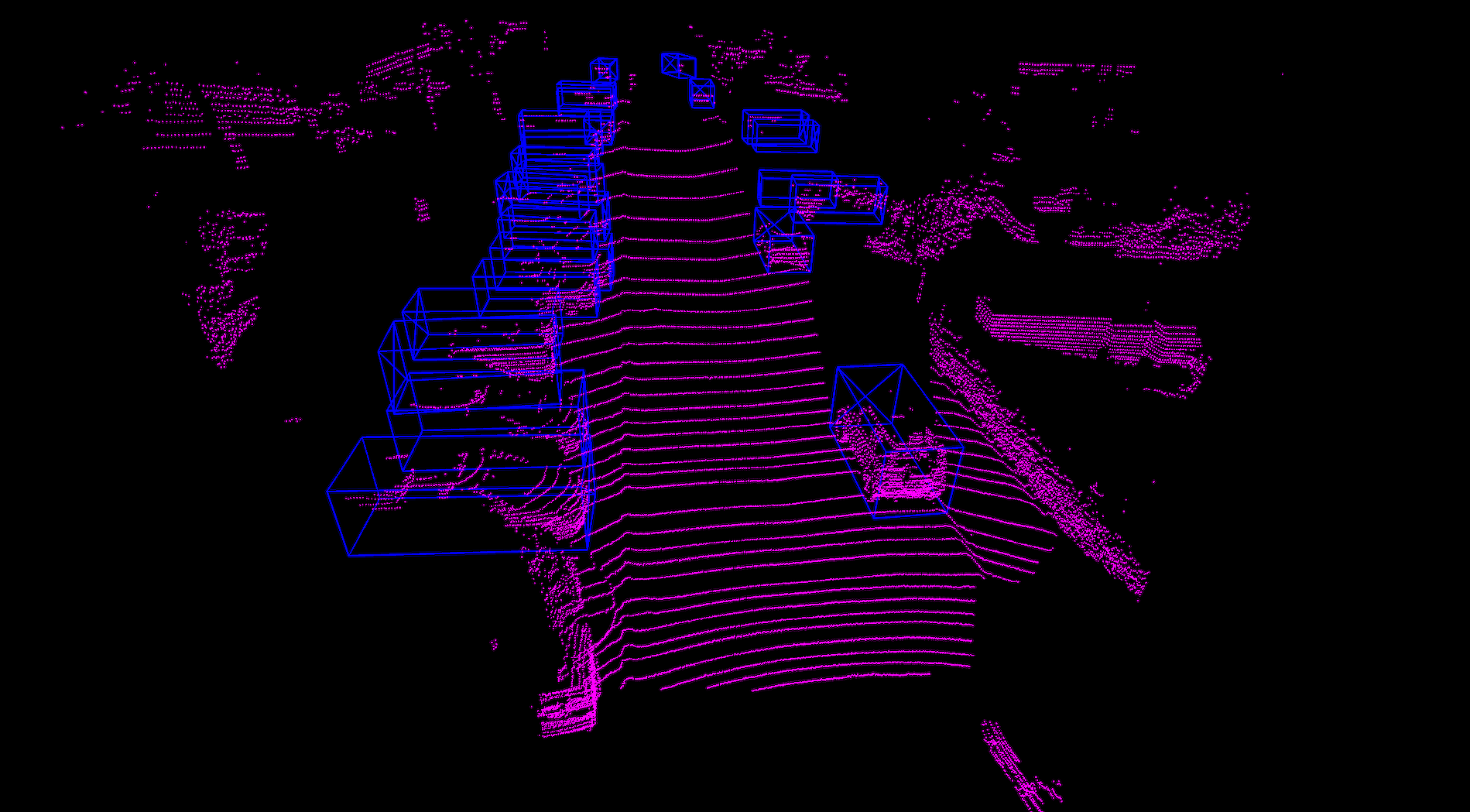}
	\end{minipage}
	\begin{minipage}{0.33\linewidth}
		\centering
		\includegraphics[width=0.9\linewidth]{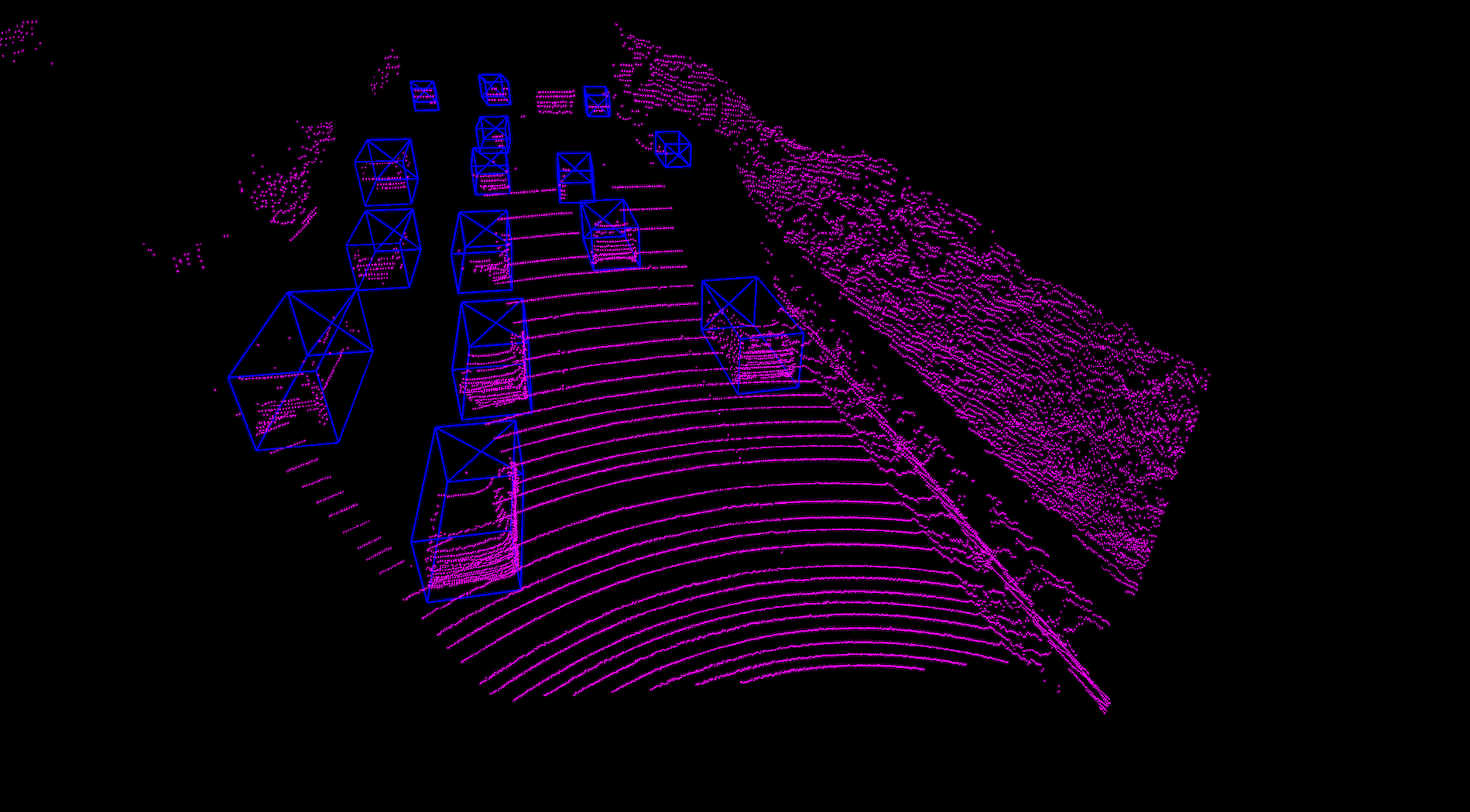}
	\end{minipage}
	\begin{minipage}{0.33\linewidth}
		\centering
		\includegraphics[width=0.9\linewidth]{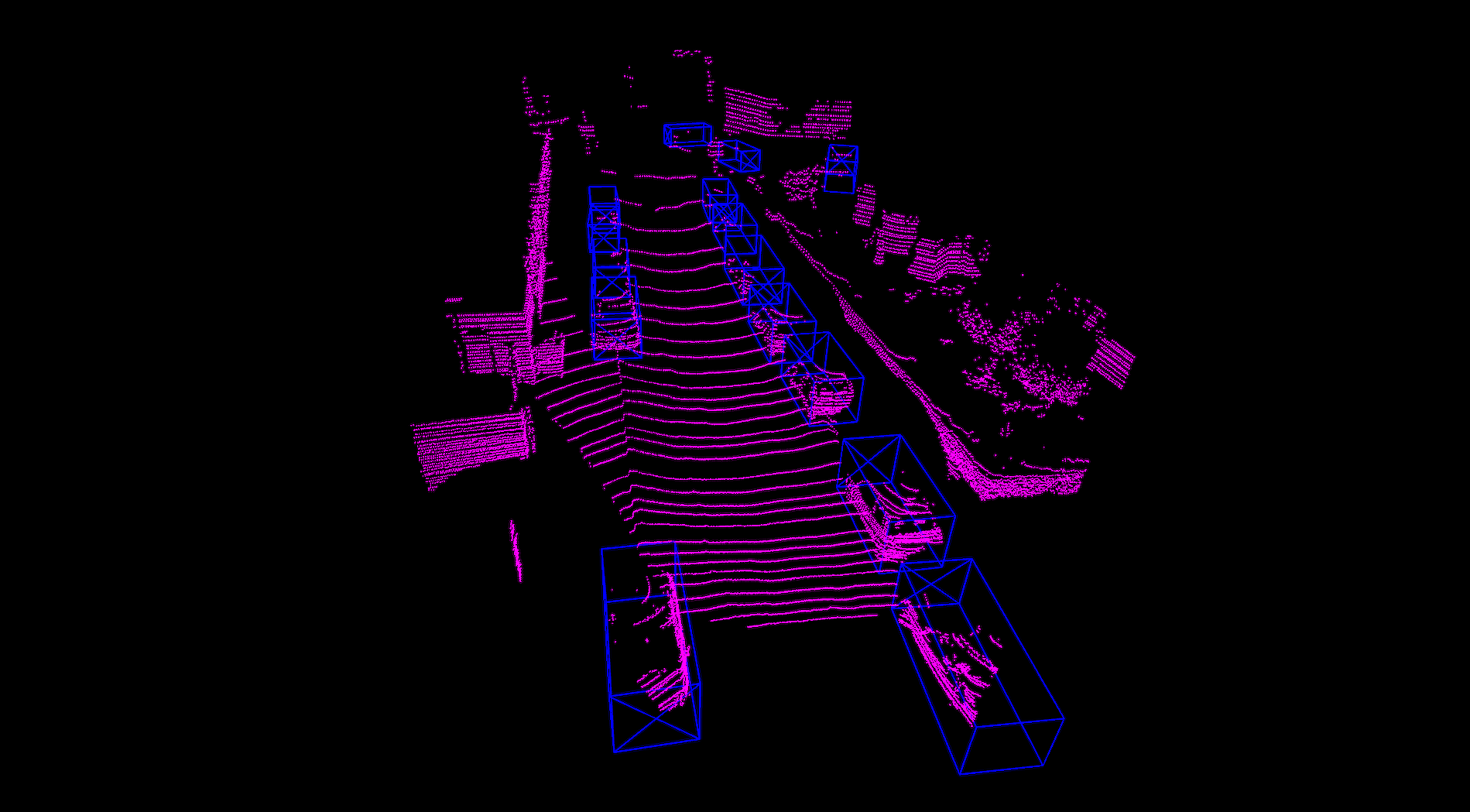}
	\end{minipage}
	\begin{minipage}{0.33\linewidth}
		\centering
		\includegraphics[width=0.9\linewidth]{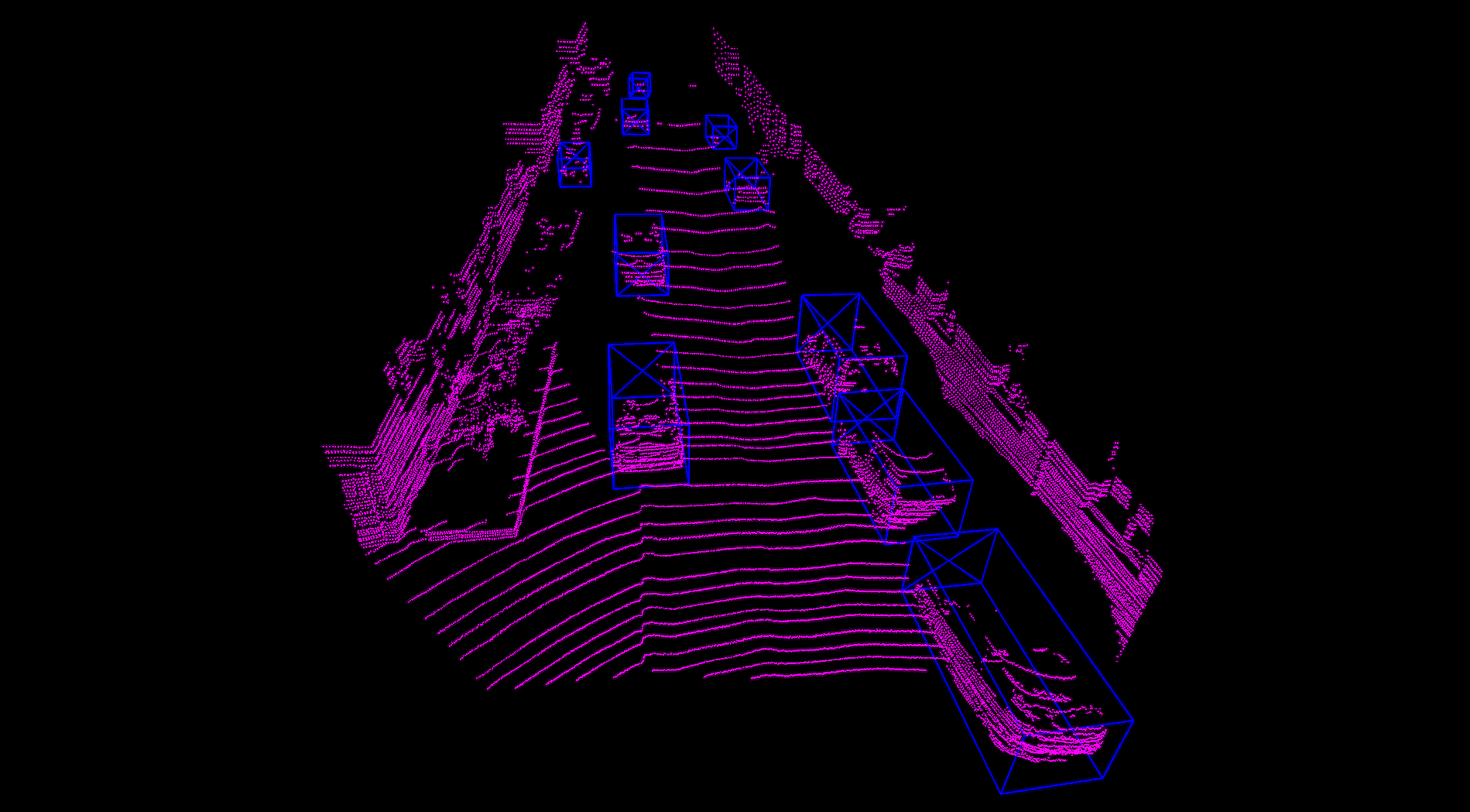}
	\end{minipage}
	\begin{minipage}{0.33\linewidth}
		\centering
		\includegraphics[width=0.9\linewidth]{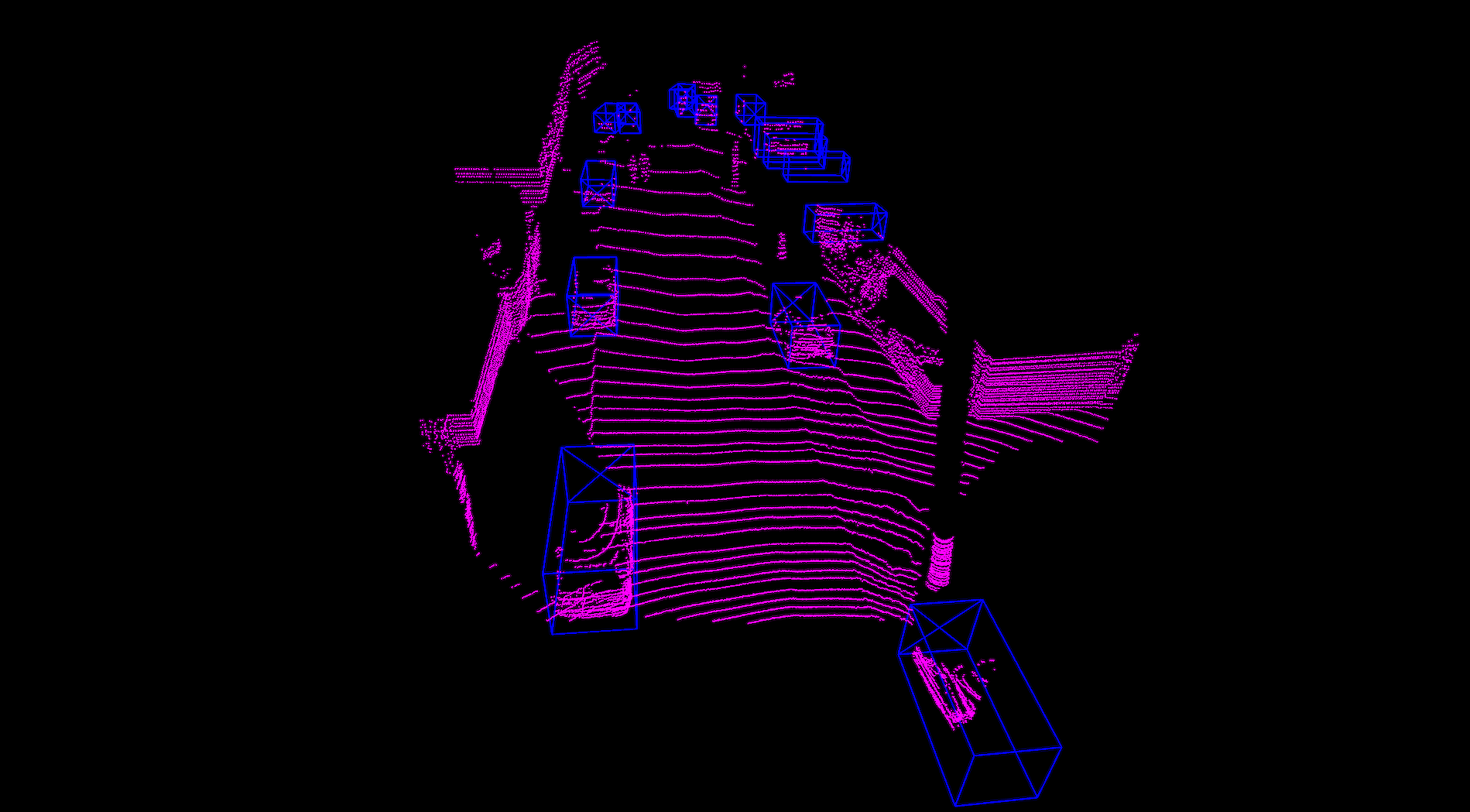}
	\end{minipage}
    \begin{minipage}{0.33\linewidth}
		\centering
		\includegraphics[width=0.9\linewidth]{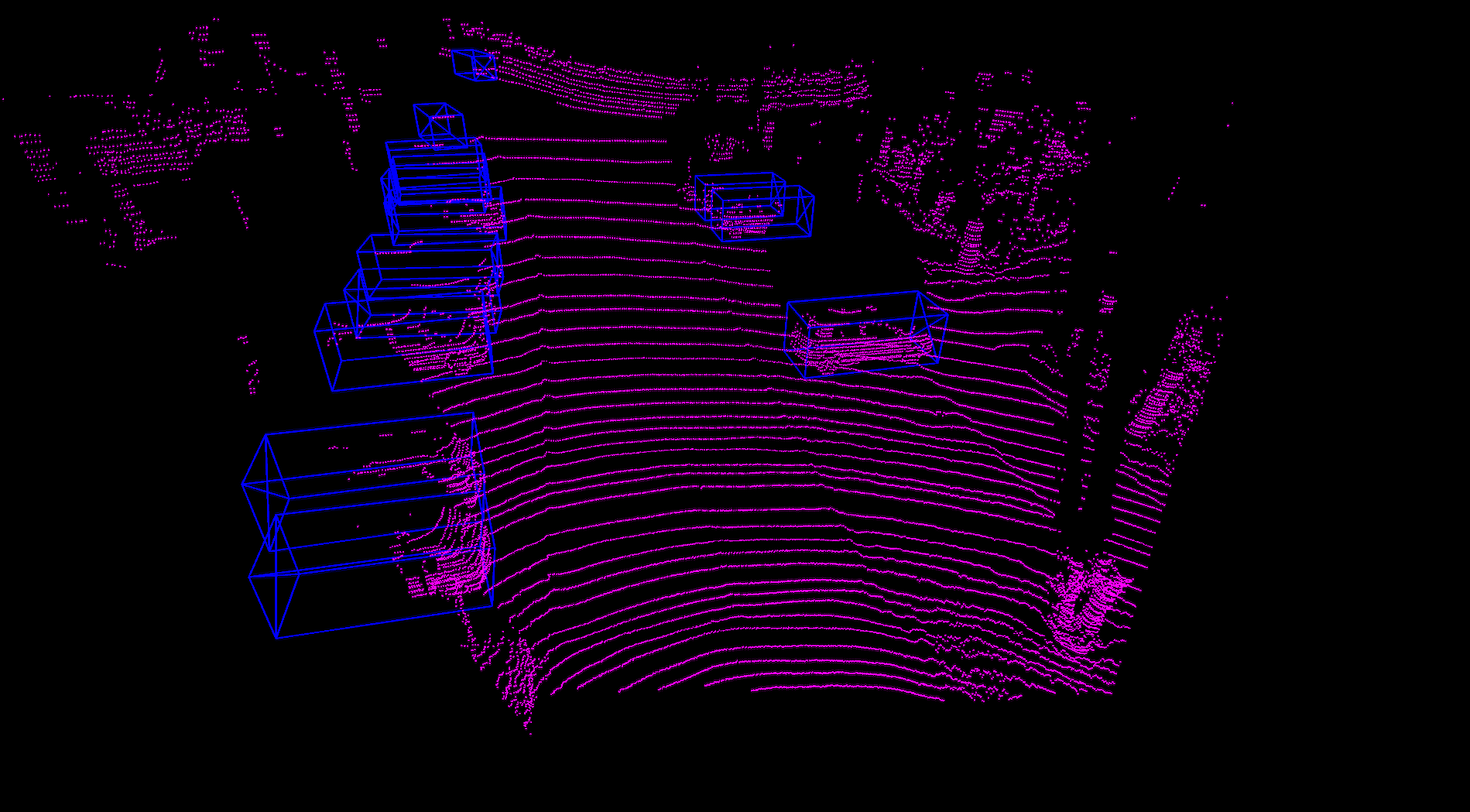}
	\end{minipage}
	\caption{Qualitative results of Waymo-to-KITTI cross-domain scenario. We visualize the detection results in the target domain (KITTI).}
	\label{fig:kitti_wo_gt}
\end{figure*}

\begin{figure*}[h]
\vspace{-0.30cm}
    \begin{minipage}{0.33\linewidth}
		\centering
		\includegraphics[width=0.9\linewidth]{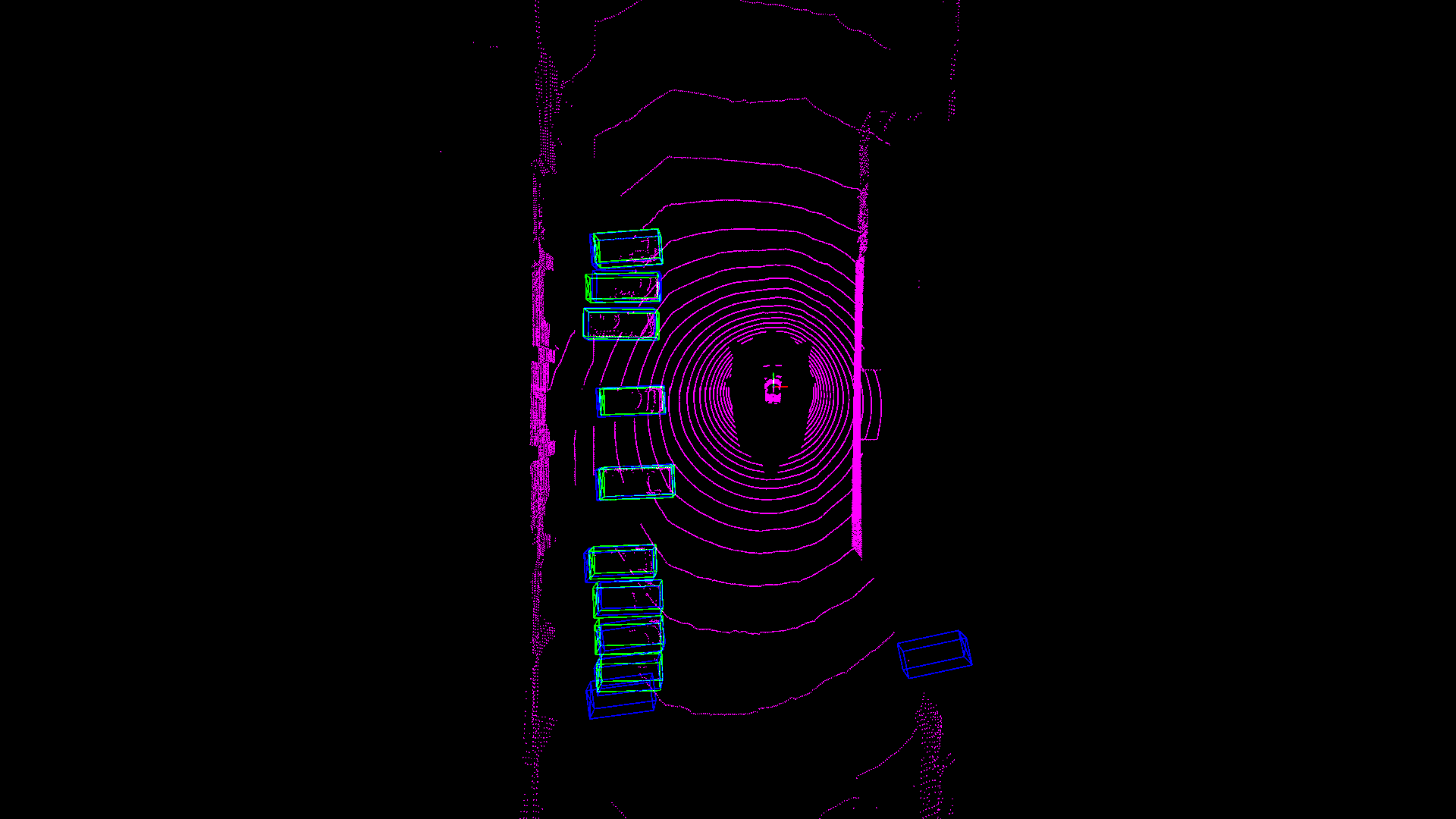}
	\end{minipage}
	\begin{minipage}{0.33\linewidth}
		\centering
		\includegraphics[width=0.9\linewidth]{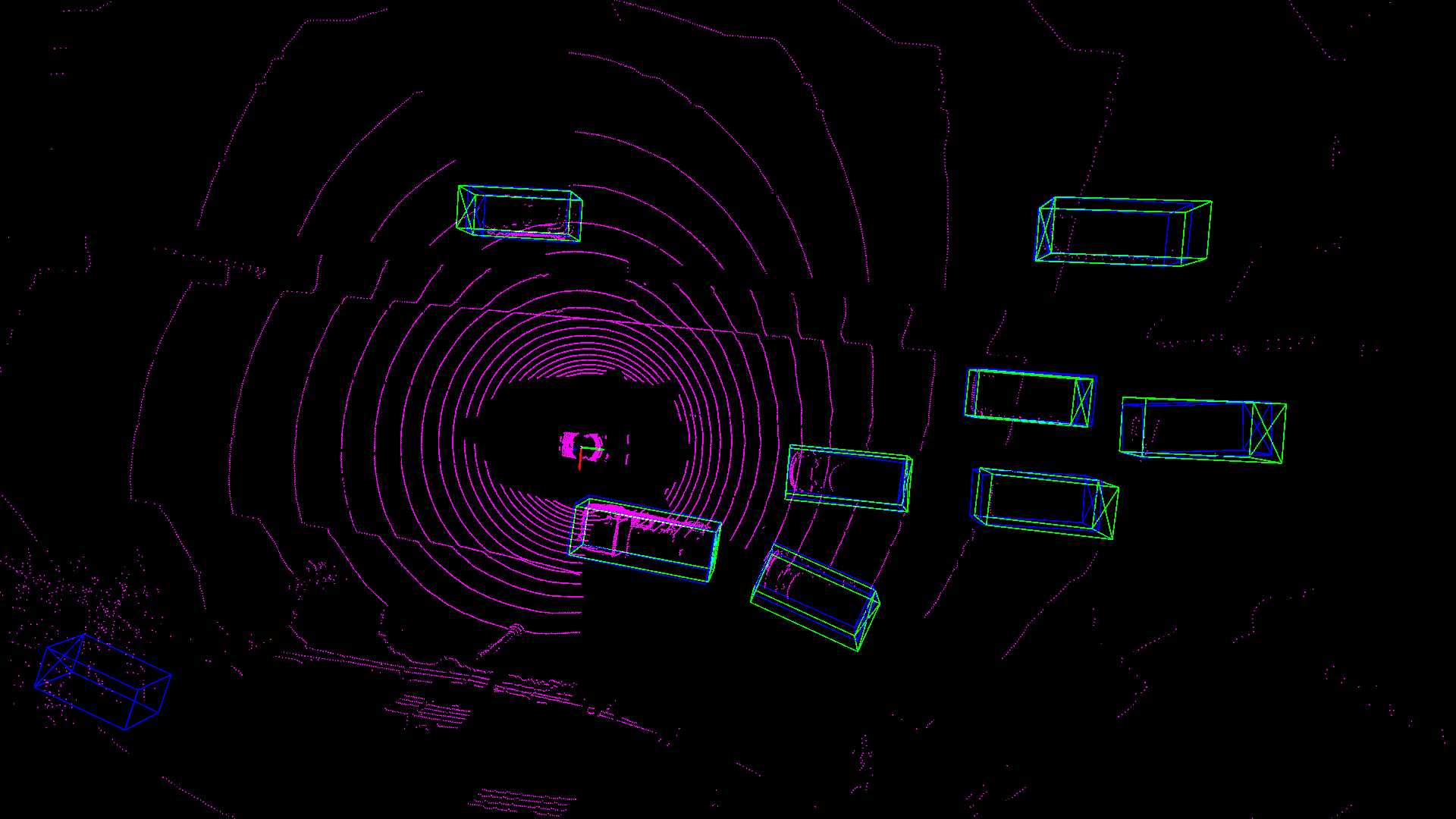}
	\end{minipage}
	\begin{minipage}{0.33\linewidth}
		\centering
		\includegraphics[width=0.9\linewidth]{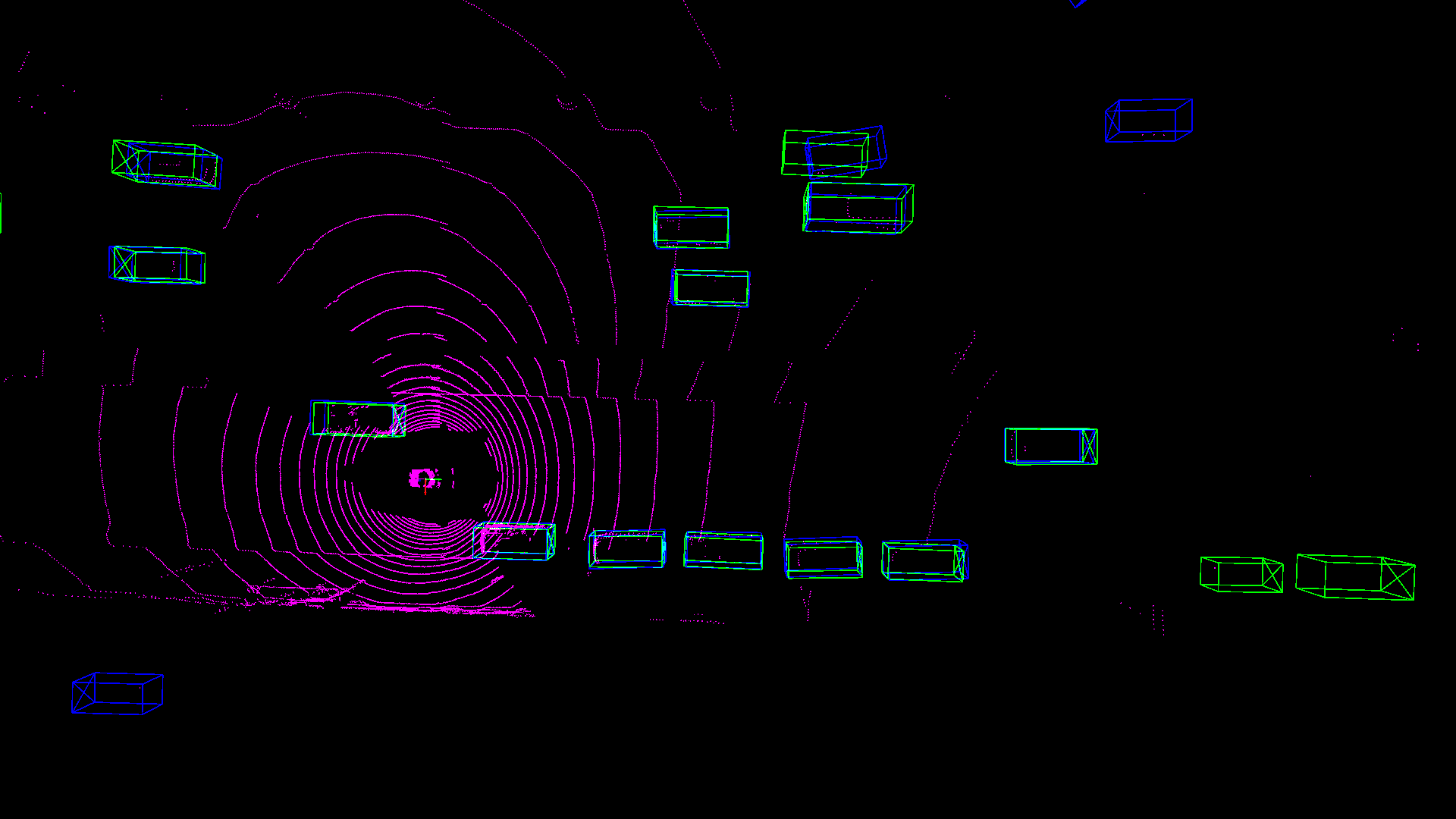}
	\end{minipage}
	\begin{minipage}{0.33\linewidth}
		\centering
		\includegraphics[width=0.9\linewidth]{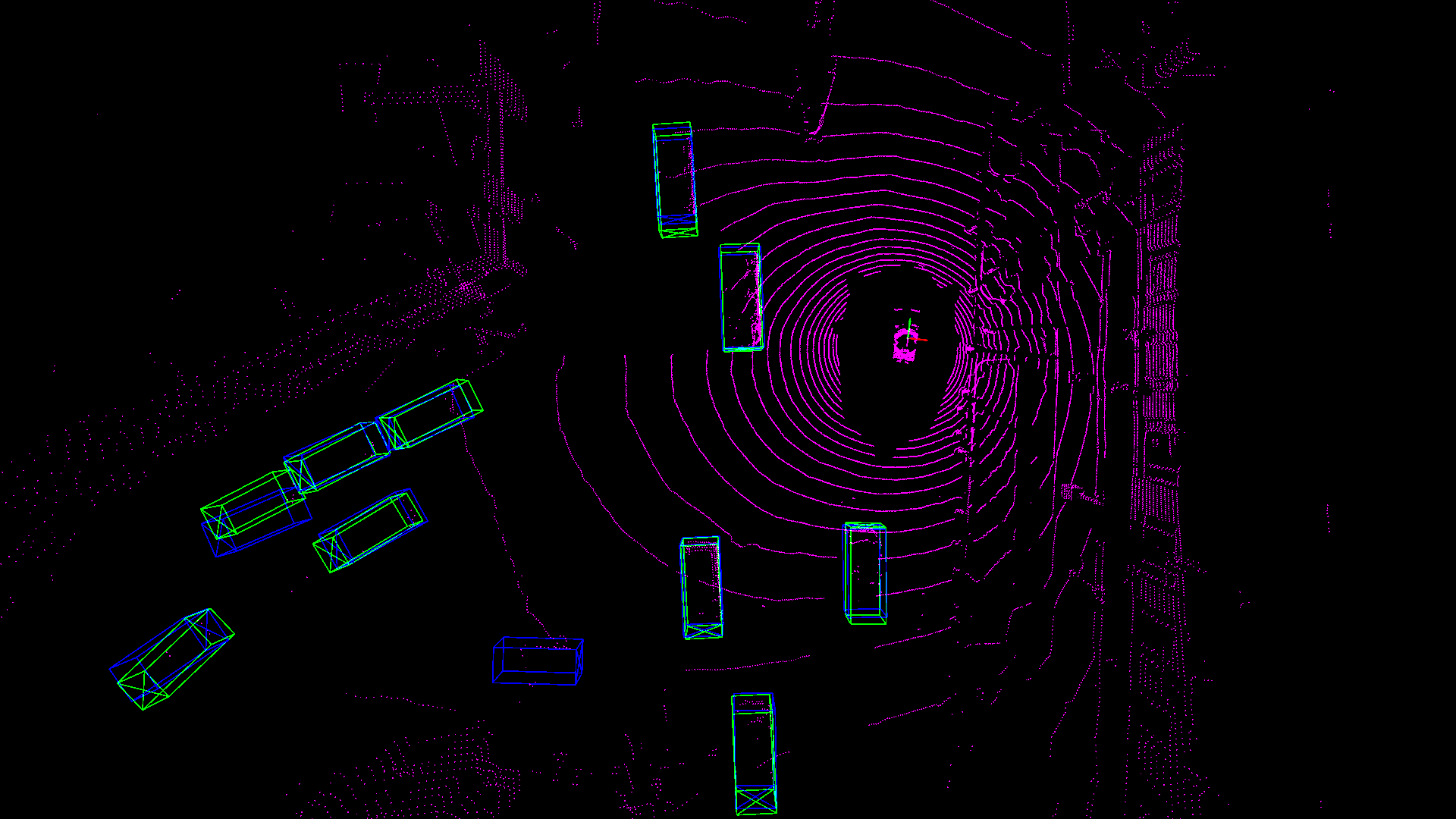}
	\end{minipage}
	\begin{minipage}{0.33\linewidth}
		\centering
		\includegraphics[width=0.9\linewidth]{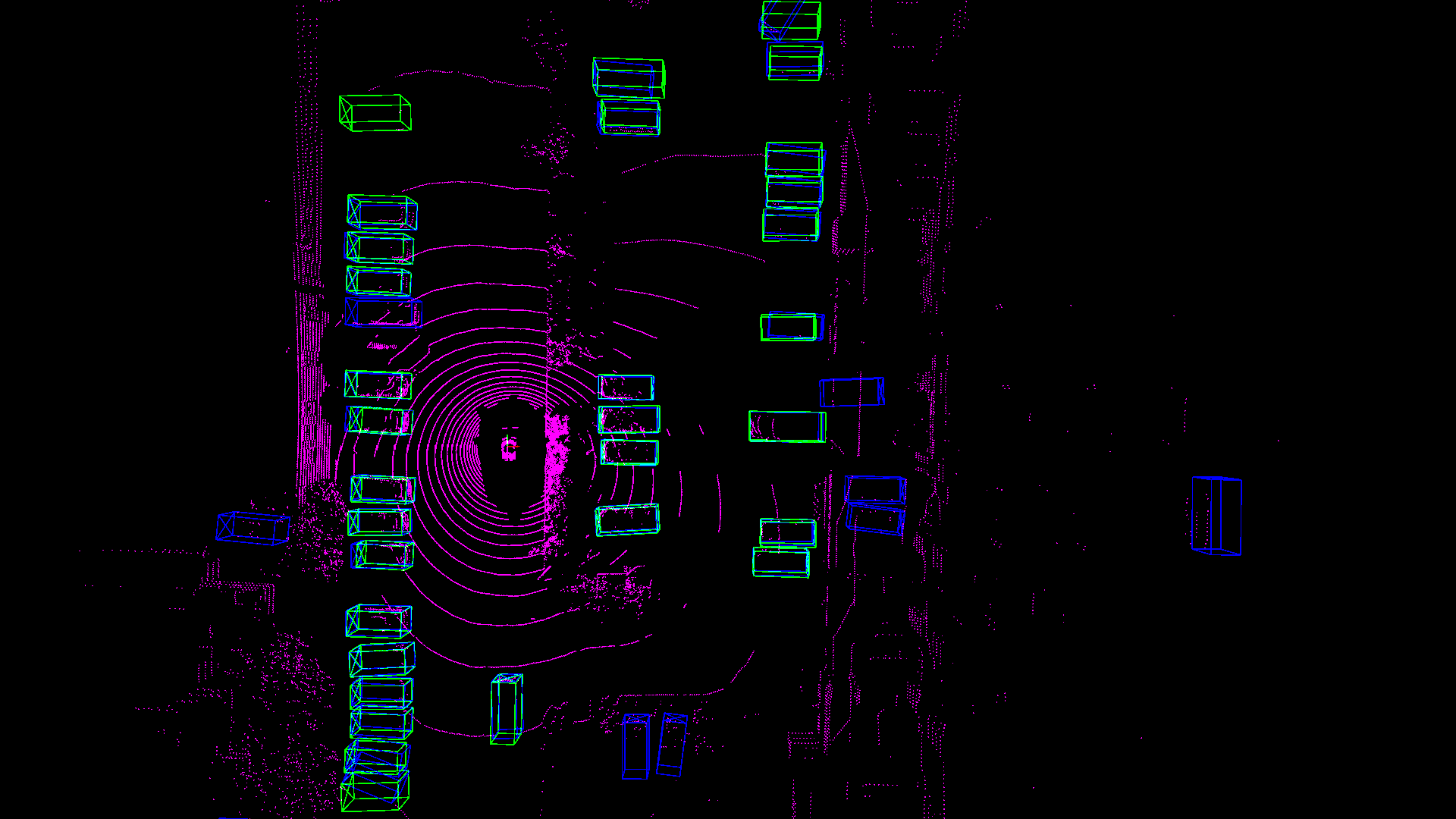}
	\end{minipage}
    \begin{minipage}{0.33\linewidth}
		\centering
		\includegraphics[width=0.9\linewidth]{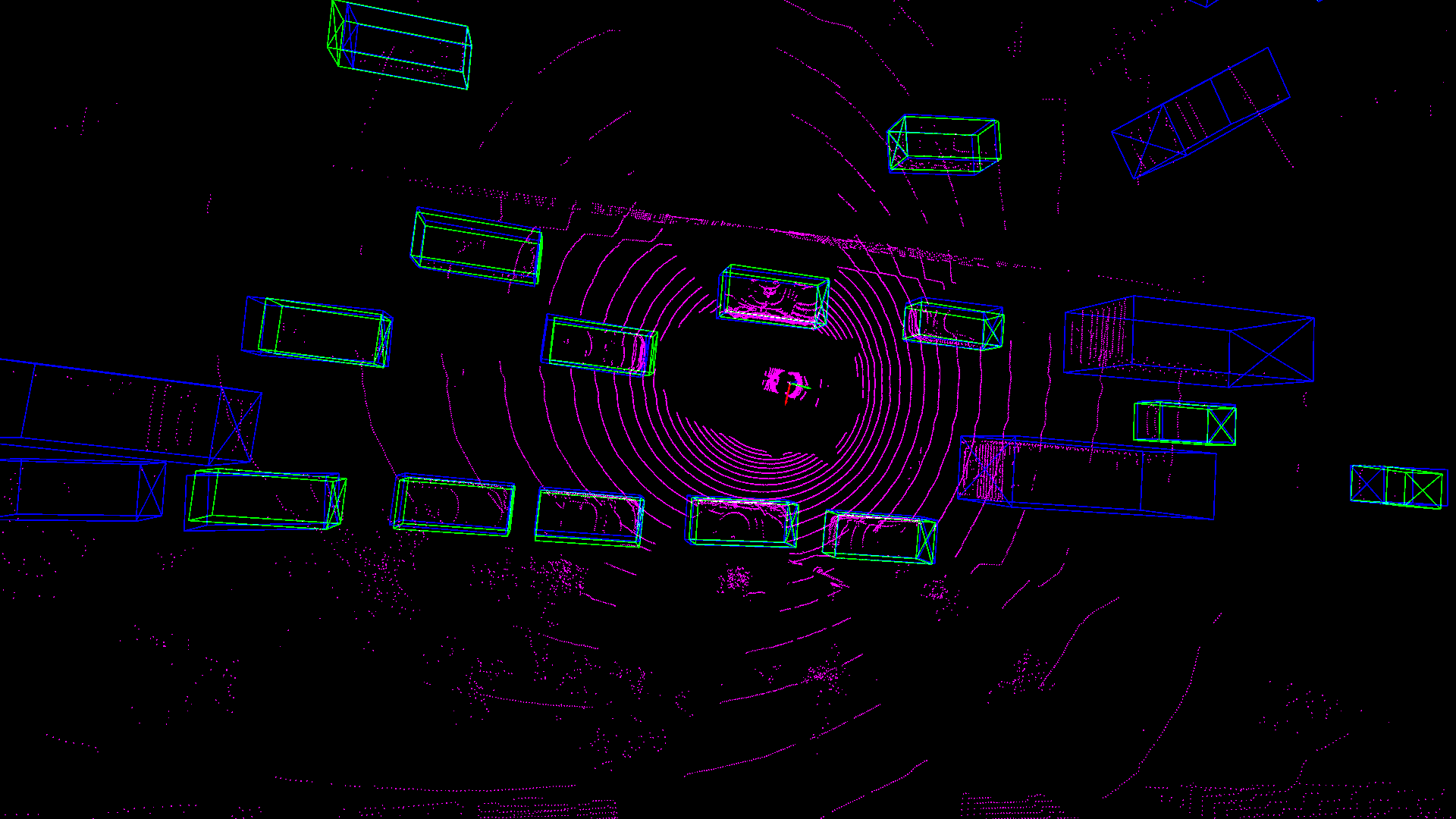}
	\end{minipage}
	\caption{Qualitative results of Waymo-to-nuScenes cross-domain scenario. We visualize the detection results in the target domain (nuScenes).}
	\label{fig:nus_with_gt}
\end{figure*}

To illustrate that Bi3D effectively samples target-like source data, we first visualize the scene-level features of source and target domains using t-SNE~\cite{van2008visualizing}. As shown in Fig. \ref{fig:source_tsne}, the selected source data distribute around the domain boundaries between the source and target domains, meaning that we sample target-domain-like source data. Besides, the visualization of instance-level features from the target domain is shown in Fig. \ref{fig:target_tsne}, and we can observe that the diverse target data are selected using Bi3D.

\subsection{Visualization}

To better verify the effectiveness of our Bi3D, we finally provide some visualizations. Fig. \ref{fig:kitti_with_gt} and Fig. \ref{fig:kitti_wo_gt} show qualitative results of Waymo-to-KITTI cross-domain scenario equipped with PV-RCNN. Fig. \ref{fig:nus_with_gt} shows the qualitative results of Waymo-to-nuScenes cross-domain scenario. It can be seen that our method can predict high-quality 3D bounding boxes. Besides, due to the differences in the taxonomy of different datasets (\eg, in the waymo dataset, cars, trucks and buses are annotated as 'Vehicle' and it is quite different from nuScenes dataset, which only annotated cars as 'car'), we can observe that the model detects trucks and buses in nuScenes, which will reduce the detection accuracy.





\end{document}